\pdfoutput=1
\documentclass[11pt]{article}

\usepackage{acl}
\usepackage{enumitem}
\usepackage{booktabs} 
\usepackage{tablefootnote}
\usepackage{tcolorbox}
\usepackage{float}
\usepackage{times}
\usepackage{latexsym}
\usepackage{makecell}
\usepackage{amsmath} 
\usepackage{amssymb}
\usepackage{bm}
\usepackage{dsfont}
\usepackage{multirow}
\usepackage{url}
\usepackage{comment}
\usepackage{bbm}
\usepackage{lipsum}
\usepackage{arydshln}
\usepackage{CJKutf8}
\usepackage{subcaption}
\usepackage{hyperref}
\usepackage{zref-xr, zref-user}
\usepackage{fontawesome}

\usepackage[T1]{fontenc}
\usepackage[utf8]{inputenc}
\usepackage{microtype}
\usepackage{inconsolata}
\usepackage{graphicx}

\title{Neuron Empirical Gradient: \\Discovering and Quantifying Neurons' Global Linear Controllability}

\author{%
    Xin Zhao \qquad\quad Zehui Jiang \\
    The University of Tokyo \\
    \texttt{\{xzhao,zjiang\}@tkl.iis.u-tokyo.ac.jp} \\\And
    Naoki Yoshinaga \\
    Institute of Industrial Science, \\
    The University of Tokyo \\
    \texttt{ynaga@iis.u-tokyo.ac.jp}
}

\newcommand{\shin}[1]{\textcolor{black}{#1}}
\newcommand{\cmr}[1]{\textcolor{black}{#1}}

\begin{document}

\maketitle

\begin{abstract}
% \shin{Although neurons in the feed-forward layers of pre-trained language models (PLMs) can store knowledge, the lack of quantitative analysis limits our understanding of how neuron activations influence model output conveying knowledge. Uncovering these relationships is crucial for improving model interpretability and controllability.}
While feed-forward neurons in pre-trained language models (PLMs) can encode knowledge, past research targeted a small subset of neurons that heavily influence outputs.
This leaves the broader role of neuron activations unclear, limiting progress in areas like knowledge editing.
We uncover a global linear relationship between neuron activations and outputs using neuron interventions on a knowledge probing dataset.
The gradient of this linear relationship, which we call \textbf{the neuron empirical gradient \shin{(NEG)}}, captures how changes in activations affect predictions.
To compute NEG efficiently, we \shin{propose \textbf{NeurGrad}, enabling large-scale analysis of neuron behavior in PLMs.}
We also show that NEG effectively captures language skills across diverse prompts through skill neuron probing. 
Experiments on \textbf{MCEval8k}, a multi-genre multiple-choice knowledge benchmark, support NEG's ability to represent model knowledge. 
\cmr{Further analysis highlights the key properties of NEG-based skill representation: efficiency, robustness, flexibility, and interdependency.}
The code and data are released.

{\centering
 \faGithub\,\href{https://github.com/xzhao-tkl/NEG}{\textcolor{black}{\texttt{xzhao-tkl/NEG}}}\,\,\,\,\,\includegraphics[height=1em]{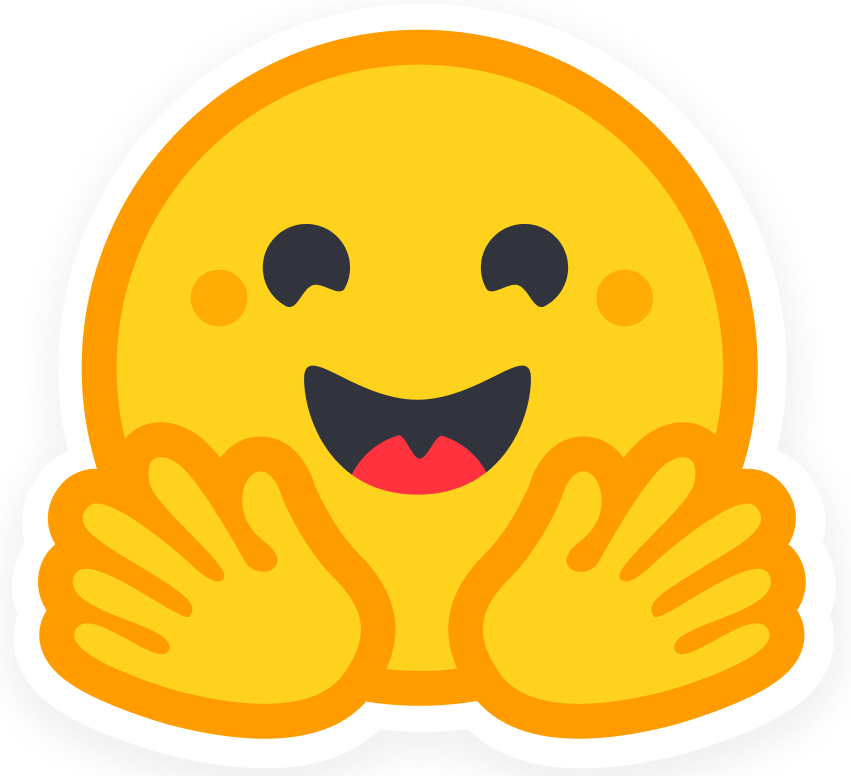}\,\href{https://huggingface.co/datasets/iszhaoxin/MCEval8K}{\textcolor{black}{\texttt{iszhaoxin/MCEval8K}}}\par}
\end{abstract}

\section{Introduction}
% Although Transformer~\citep{vaswani2017attention}-based language models (LMs) benefit from large-scale pre-training,
% the pre-trained LMs (PLMs) suffer from hallucination, where models generate incorrect knowledge. This issue makes it important to understand the mechanism by which PLMs store knowledge within their parameters~\citep{dai-etal-2022-knowledge,niu2024what,wang-etal-2024-knowledge-mechanisms,wang-etal-2022-finding-skill}.

\shin{Pre-trained language models (PLMs) based on Transformer architecture~\cite{vaswani2017attention}
% exhibit a strong ability to 
effectively encode human knowledge, prompting efforts to understand their inner workings.}
% In Transformer~\citep{vaswani2017attention}-based language models (LMs), feed-forward (FF) layers serve as key-value memory~\cite{geva-etal-2021-transformer}, with neurons able to retrieve knowledge. 
% Previous work reveals that specific facts correlate with a limited number of neurons~\citep{dai-etal-2022-knowledge, yu-ananiadou-2024-neuron}, and even a few neurons own the abilities to perform various language skills associated with diverse prompts~\citep{wang-etal-2022-finding-skill, tan-etal-2024-neuron}. 
\shin{While previous studies have shown that the feed-forward (FF) neurons play key roles in encoding factual knowledge~\citep{dai-etal-2022-knowledge, yu-ananiadou-2024-neuron} and general language skills~\citep{wang-etal-2022-finding-skill, tan-etal-2024-neuron}, they face two main challenges.
First, current methods mostly rank neurons by importance without measuring the link between neuron activations and model outputs \citep{dai-etal-2022-knowledge, meng2022locating, yu-ananiadou-2024-neuron}, 
%yet they lack a direct and global quantitative measurement between neuron activations and model output, 
limiting use cases
% their applicability to neuron-level adjustment applications, 
like knowledge editing~\cite{zhang-etal-2024-knowledge-editing}.
Second, these methods are costly, involving repeated
% either multiple 
activation modifications~\citep{dai-etal-2022-knowledge, meng2022locating, goldowskydill2023localizingmodelbehaviorpath} or extensive tensor operations~\citep{yu-ananiadou-2024-neuron}, making them inefficient for analyzing all neurons on large models.}
% diverse prompts.}
% thereby hindering a global understanding of their roles.}
% Although these studies reveal the role of neurons in representing factual knowledge and language skills, the lack of quantitative analysis limits our understanding of how to control model behavior at the neuron level.

\begin{figure}[t]
    \centering
    \includegraphics[width=\linewidth]{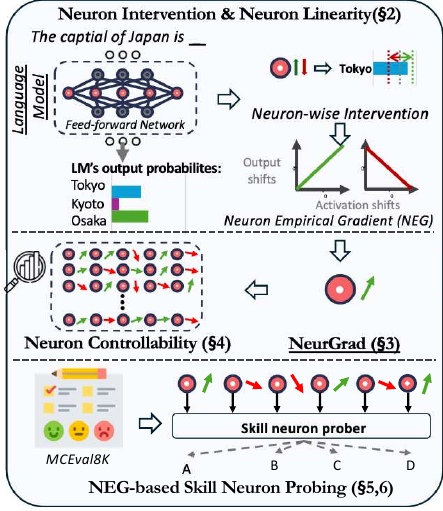}
    \caption{Our contributions: i) revealing linearity between activation and output shifts, ii) proposing NeurGrad, an efficient method to quantify it, iii) confirming that NEGs capture language skills on MCEval8K.}
    \label{fig:overview}
\end{figure}

This study begins by quantitatively analyzing \textbf{how neuron activations influence model outputs (RQ1)} (Figrue~\ref{fig:overview}). Using MyriadLAMA~\citep{zhao-etal-2024-matters}, a factual knowledge probing dataset, on seven PLMs, including large language models (LLMs) like Llama2-70B (\S~\ref{sec:neuron_analysis}),
we gradually modify random neuron activations and observe the changes in the token probabilities for correct knowledge (hereafter, \textit{output shifts}).
% using MyriadLAMA~\citep{zhao-etal-2024-matters}, 
% a factual knowledge probing dataset on nine PLMs, including large language models (LLMs) like Llama2-70B (\S~\ref{sec:neuron_analysis}).
Notably, we find that within a certain range, shifts in neuron activations (hereafter, \textit{activation shifts})
% in neuron activations (hereafter, \textbf{activation shifts}) 
have a linear impact on output shifts.
% a linear relationship with the output shift. 
We define and quantify the gradient of this linear relationship as \textit{neuron empirical gradient \shin{(NEG)}}, enabling quantitative neuron analysis.

% Furthermore, we find neurons vary in the shifting direction of output probabilities when increasing neuron activations.
% We term this property of neurons as \textit{polarity}, which we use to classify neurons as either \textit{positive} or \textit{negative}.
% We also find that neurons differ in the direction they shift output probabilities as their activations increase — a property we call \textit{polarity}, which allows us to classify neurons as positive or negative.
% Our evaluation of 9 PLMs, including Llama2-70B, confirms that neurons generally exhibit linearity.  
% We term the gradient of the linear relationship between a specific neuron with a token in response to a prompt as the \textbf{neuron empirical gradient \shin{(NEG)}}.

% While NEG quantifies a neuron's importance and direction in shaping PLM outputs, their calculation is costly due to variability across prompts, neurons, and target tokens.
% To facilitate quantitative neuron analysis, we thus propose NeurGrad, an efficient method for estimating NEG, and validate its performance on the MyriadLAMA dataset (\S~\ref{sec:neurgrad}).
% Our results on the above six, diverse PLMs show that NeurGrad outperforms baseline methods in both efficiency and precision.

\shin{Next, we explore \textbf{whether shifting neuron activations can precisely control PLMs' output probabilities (RQ2)}. 
Since computing NEGs requires costly inference, we propose \textit{NeurGrad} (\S~\ref{sec:neurgrad}), an efficient method for estimating a single neuron's NEG\@. It builds on the empirical finding that computational gradients (\S~\ref{sec:neurgrad-proposal}) strongly correlate with NEG magnitudes, though less so with directions. 
Using MyriadLAMA, we validate NeurGrad's performance against ground-truth NEGs, showing it outperforms existing neuron-ranking methods~\citep{dai-etal-2022-knowledge,yu-ananiadou-2024-neuron}. 
With NeurGrad, we further examine multi-neuron control, finding that NEGs can accumulate across neurons (\S~\ref{sec:understand-neuron-controllability}), while the effect weakens as more neurons are involved or larger activation shifts are involved.}
% This reveals the reason behind the neuron linearity: the local approximation to the model's differentiable behavior.
% propose two hypotheses to advance our understanding of knowledge storage and neuron behavior: the \textbf{dynamic knowledge store hypothesis}, which explains how knowledge is dynamically stored in PLMs, and the \textbf{local linear approximation hypothesis}, which accounts for neuron linearity.
% \shin{NEG reveals the \textbf{neuron's controllability}: allowing precise adjustments to PLMs' output probabilities through targeted activation shifts. 
% \textbf{We further investigate neuron controllability in greater depth, such as how NEG reveals the underlying mechanism of knowledge representation.} 
% However, such analysis requires the acquisition of NEG of all neurons, and calculating NEG through neuron intervention is costly.
% due to its variability across prompts, neurons, and target tokens.
% We thus propose NeurGrad, an efficient method for estimating NEG, and validate its performance in estimating NEG on MyriadLAMA (\S~\ref{sec:neurgrad}).
% Our results on the above six, diverse PLMs show that NeurGrad outperforms baseline methods in both efficiency and precision.

\shin{Finally, we examine \textbf{whether NEGs can capture general language skills 
% across diverse prompts 
beyond factual knowledge (RQ3)}.}
% \shin{While NEG enables the output modification for a specific prompt and target token, modifying model behavior based on knowledge requires handling diverse prompts that convey the same knowledge type (hereafter, \textbf{language skill}), such as sentiment classification.
% Our study aims to establish the foundation for such modification by asking: \textbf{whether can NEG represent language skills?}
To explore this, we perform skill neuron probing~\citep{wang-etal-2022-finding-skill, song-etal-2024-large} 
%that identifies neurons associated with language skills 
using NEGs.
% as inputs.While prior studies conducted skill neuron probing, they focus on using neuron activations to build probes, leaving the representational ability of NEG unexamined~\citep{wang-etal-2022-finding-skill, song-etal-2024-large}. 
As the previous studies on skill neuron probing focus on limited language skills, we introduce \textit{MCEval8K}, a benchmark covering six genres and 22 tasks for broad LLM evaluation.
Our results show that NEGs can represent a wide range of language skills. 
\cmr{Furthermore, our in-depth analysis highlights the properties of NEG-based skill neurons, including the efficiency in representing language skills, robustness when facing diverse contexts, substitutability of skills being represented by different neurons, and interdependency between neurons in representing skills.}

Our contributions (Figrue~\ref{fig:overview}) are as follows:\begin{itemize}
\setlength{\itemsep}{-1pt}
\item We confirm that activation and output shifts are linearly correlated within a certain range through neuron intervention, defined as \textbf{neuron empirical gradient (NEG)} (\S~\ref{sec:neuron_analysis}).
\item \shin{We introduce \textbf{NeurGrad}, an efficient method to NEG (\S~\ref{sec:neurgrad}), and conduct in-depth analysis about neuron controllability (\S~\ref{sec:understand-neuron-controllability}).} 
\item \shin{We show that NEGs can represent language skills through \textbf{skill neuron probing} (\S~\ref{sec:skill-neuron-probing-task})}; \cmr{skill neurons exhibit efficiency, robustness, inclusivity, and interdependency (\S~\ref{sec:skill-neuron-property}).} 
% empirical gradients capture diverse language skills, 
% uncovering properties of knowledge representation.
\item We develop \textbf{MCEval8K}, a multiple-choice benchmark covering six genres and 22 language understanding tasks (\S~\ref{sec:mceval8k}, \S~\ref{sec:appendix-mceval8k}). 

\end{itemize}

\section{Neuron Linearity to Model Output}
\label{sec:neuron_analysis}
% This section quantitatively analyzes how neurons in PLMs' FF layers influence model generations. 
This section empirically answers how neurons in PLMs' FF layers influence model outputs. We observe the resulting change in output tokens' probabilities for fine-grained neuron-level interventions.

% Using factual knowledge probing as the target task, we perform neuron-wise intervention by adjusting neuron activations for the same prompt and observing the resulting change in output tokens' probabilities.
% Experiments on six PLMs, including both masked and casual models, indicate that neurons exhibit linearity and polarity. 
% The signed slope of the linear relationship between activation shifts and output probabilities is termed the empirical gradient.

\subsection{Neuron-level Intervention Experiments}
\label{ssec:experiment}
\paragraph{Models.}
To ensure the generality of our findings, we evaluate both masked and causal LMs with varying sizes and learning strategies.
\cmr{For masked LMs, we use two BERT~\cite{devlin-etal-2019-bert} models, BERT$_\mathrm{base}$ and BERT$_\mathrm{large}$, and have them predict masked tokens.
For causal LMs, we examine five LLMs with different model sizes and language families, including Llama2~\citep{touvron2023llama2openfoundation} (7B, 70B), Llama3.1 (8B), Llama3.2 (3B)~\citep{grattafiori2024llama3herdmodels}, and Qwen2.5 (7B)~\citep{qwen2025qwen25technicalreport}. 
All of these models are instruction-tuned.}
Following \citet{zhao-etal-2024-matters}, we use a zero-shot prompt to generate single-token answers. 
See \S~\ref{sec:appendix-mode_cards} for details.

\paragraph{Dataset.}
We use MyriadLAMA~\citep{zhao-etal-2024-matters}, a multi-prompt knowledge probing dataset for neuron intervention.  
Its diverse prompts help reduce bias from specific phrasing.
We focus on single-token probing, where the target answer is a single token. 
For each PLM, we randomly sample 1000 prompts from MyriadLAMA, where the model predicts the token representing the correct answer. 
Due to tokenizer and knowledge differences, the sampled prompts vary across PLMs.

\paragraph{Neuron intervention.}
We shift activations in the range of [-10, 10]\footnote{The range mostly covers the distribution of activations; see \S~\ref{sec:appendix-activation-distribution} for details.} with the step size of 0.2 to track changes in target token probabilities.
\cmr{For example, consider the following prompt:}
\begin{quote}
Predict the \texttt{[MASK]} in each sentence in one word.\\
Q: \texttt{[MASK]} is the capital of Japan.\\
A: 
\end{quote}
\cmr{We modify the activations of specific neurons and observe how the probability of the target word \emph{Tokyo} changes at the final position.}
Since conducting one intervention experiment for a neuron-token-prompt combination requires 100 inferences due to the shift values, we randomly sample\footnote{The actual number of randomly sampled neurons will be given in each experiment.} neurons to make computation tractable while ensuring broad coverage.
% Computational gradient refers to the gradient computed from the computational graph through backpropagation.

% \subsection{Results and Analysis}
% \label{neuron-linearity}
% Experimental results on a few neurons show a linear relationship between activation shifts and output probabilities. We then investigate whether this linearity is a general property of neurons.

\begin{figure}[t]
    \centering
    \includegraphics[width=\linewidth]{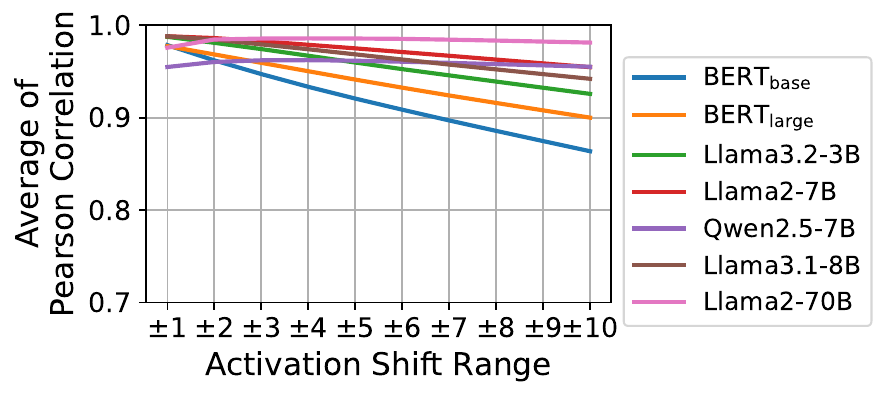}
    \caption{Average absolute correlation between activation and output shifts.} 
    \label{fig:shift_range}
\end{figure}

\smallskip\noindent\textbf{Results.}
\shin{To understand how output shifts respond to neuron activation shifts, we compute the Pearson correlation coefficient ($r$) between activation shifts and output shifts of correct tokens, using absolute $r$ values.
To reduce computation, we average $r$ over 10 randomly sampled prompts from the overall 1000 prompts, each with 1000 randomly sampled neurons at specific shift ranges (x-axis).}
Figure~\ref{fig:shift_range} shows a strong, nearly linear correlation across a wide $\pm$10 range, with stronger correlation at smaller shift ranges consistent across all models.
This suggests predictable output changes from specific activation shifts.
Since all PLMs behave similarly, we focus on BERT models and Llama2 LLMs in subsequent analysis.

% Specifically, when setting the activation shift range to $\pm$2, the Corrs. in most models are above 0.95, which we consider the threshold for indicating the linear relationship.
% }
% Our subsequent analysis uses the shift range of ±2 by default.}
% Note that Llama2 and BERT models differ significantly in gradient magnitudes, with Llama2's gradients being five orders of magnitude smaller than BERT's. See \S~\ref{sec:appendix-noise} for details.}

\subsection{Neuron Linearity}
\label{neuron-linearity}

\shin{Based on the findings above, we ask \textbf{whether neurons generally show linearity with model output.} 
We define neurons as linear if their correlation ($r$) is at least 0.95\footnote{We assume $r\ge 0.95$ to be a strong linear relationship.} within a $\pm$2 shift range, as observed in Figure~\ref{fig:shift_range}.}

Then, \shin{we quantitatively analyze neuron linearity across prompts and Transformer layers by measuring the ratio of linear neurons from 1000 prompts and 100 \cmr{randomly sampled} neurons.
\footnote{We only chose 200 prompts and 100 neurons for Llama2-70B due to the large model size.} 
\cmr{The ratios of linear neurons are reported in Table~\ref{tab:neuron_ratios}, showing that most neurons in LLMs exhibit linearity.}
This linearity is widespread across layers and prompts (\S~\ref{sec:appendix-neurongen}).
We also define \textbf{polarity} as follows: neurons are \textit{positive} if increasing activations boost target token probabilities, and \textit{negative} otherwise.} The analysis of neuron polarity is deferred to \S~\ref{sec:neuron-distribution}.

\begin{table}[t]
    \centering
    \small
    \begin{tabular}{lccc}
        \toprule
        & \textbf{Linear} & \textbf{Positive} & \textbf{Negative}  \\
        \midrule
        BERT$_\mathrm{base}$  & .9201 & .4981 & .5004  \\
        BERT$_\mathrm{large}$ & .9209 & .4938 & .4981 \\
        \midrule
        Llama3.2-3B           & .9483 & .5000 & .5000 \\
        Llama2-7B             & .9659 & .5157 & .4843 \\
        Qwen2.5-7B            & .6549 & .5116 & .4884 \\
        Llama3.1-8B           & .9540 & .5048 & .4952\\
        Llama2-70B            & .9208 & .4962 & .5039 \\
        \bottomrule
    \end{tabular}
    \caption{Ratios of linear neurons and neuron polarity over 1000 prompts with 100 neurons.}
    \label{tab:neuron_ratios}
\end{table}

\paragraph{Neuron empirical gradient.} 
\shin{We quantify neuron linearity and polarity using the gradient of the linear relationship between activation and output shifts, termed \textit{neuron empirical gradient} (NEG). 
To compute NEGs, we fit a zero-intercept linear regression between activation shifts and output shifts acquired through neuron intervention; the regression coefficient serves as the NEG for each neuron, prompt, and token.}
% \shin{NEG implies the potential of \textbf{neurons' controllability}: precisely adjusting PLMs' output probabilities through shifting neuron activations.}

\section{NeurGrad for NEG Estimation}
\label{sec:neurgrad}

% \shin{NEG reveals the \textbf{neurons' controllability}: we can precisely adjust PLMs' output probabilities through targeted activation shifts within a specific range.}
\shin{Efficient and accurate computation of NEG is essential for quantitative neuron-level interpretability in PLMs. However, 
% quantifying NEG through 
the neuron-wise intervention is costly.
% impractical due to the high computational cost.
While prior knowledge attribution methods estimate neurons' influence on model outputs, they either require intensive computation or only provide relative importance, without directly measuring NEG~\citep{dai-etal-2022-knowledge, geva-etal-2022-transformer, meng2022locating, yu-ananiadou-2024-neuron}.}

\subsection{NeurGrad}
\label{sec:neurgrad-proposal}
\shin{In this section, we propose NeurGrad, an accurate and efficient method for estimating NEG, to support further analysis.
This approach is based on preliminary findings using computational gradients\footnote{\zlabel{fn:cg}Computational gradient refers to the gradient computed from the computational graph through backpropagation.}
(\textbf{CG}) to approximate NEG\@.
We compute CG and ground-truth NEG values for seven PLMs, including BERT variants and instruction-tuned Llama2, using 1000 prompts and 100 neurons per prompt with a shift range of $\pm2$. 
While CG values show a low correlation with NEG directly (average $r=-0.429$), their absolute values are highly correlated (average $r=0.961$).
%We observe that the Corr between CGs' absolute values and NEGs is high, but
Additionally, both CG and neuron activations determine the sign of NEG.}
% Calculating NEG for all neurons in PLMs is computationally costly. 
% We need an efficient NEG estimation method to facilitate further analysis of neuron controllability.}
% \shin{We first evaluate the precision of using computational gradients\footnote{Computational gradient refers to the gradient computed from the computational graph through backpropagation.} (hereafter, CG)
% to estimate NEG\@.
%\shin{Specifically, we collect ground-truth NEGs from 1000 prompts with 100 neurons per prompt, with a shift range of $\pm$2 on six PLMs, BERT families and three Llama2 instruction-tuned LLMs.
% After collecting CG on these prompt-neuron pairs, the data reveal a low Corr (-0.429 on average) between the CG and NEG, but a high Corr (0.961) between their absolute values.}
% Moreover, the sign of NEG correlates with the signs of both activation and CG\@.
% Based on these findings, we propose
Based on these results, we introduce NeurGrad: 
\begin{align}
    \bar{G_E} &= \text{CG} \times \operatorname{sign}(A) ,
\end{align}
where $\bar{G_E}$, $A$, and $\operatorname{sign}(A)$ denote the estimated NEG, activation, and sign of $A$ (1 if $A>0$ and -1 if $A<0$), respectively.
% \footnote{For neurons with zero activation, we assign an NEG of zero as such cases are rare
% \shin{, as zero-activation neurons are less than 1\% in BERT and 0.3\% in LLama2 models.}} 

\begin{table*}[t]
\centering
\small
\tabcolsep 5.5pt
\begin{tabular}{lcccccccc}
\toprule
& \multicolumn{4}{c}{\textbf{Correlation ($r$)}} & \multicolumn{3}{c}{\textbf{MAE}} \\
\cmidrule(lr){2-5}
\cmidrule(lr){6-8}
& \textbf{CG} & \textbf{IG} & \textbf{LPI} & \textbf{NeurGrad} & \textbf{CG} & \textbf{IG} & \textbf{NeurGrad} \\\midrule
BERT$_\mathrm{base}$  & -.8909 & .7360 & -\protect\footnotemark
 & \textbf{.9998} & 6.1e-03 & 3.0e-03 & \textbf{2.6e-05}\\
BERT$_\mathrm{large}$ & -.9307 & .7167 & - & \textbf{.9958} & 4.6e-03 & 2.1e-03 & \textbf{1.9e-04}\\
\midrule
% BERT$_\mathrm{wwm}$   & -.8914 & .8584 & - & \textbf{.9989} & 4.5e-03 & 2.2e-03 & - & \textbf{2.3e-05}\\
Llama3.2-3B           & -.1281  & .5854 & - & \textbf{.8094} & 3.5e-06 & 1.8e-06 & \textbf{1.6e-06}\\
Llama2-7B             & .3023  & .5377 & .6469 & \textbf{.8135} & 1.7e-06 & \textbf{1.2e-06} & 1.3e-06\\
Qwen2-7B             & .1043  & .2939 & - & \textbf{.5862} & 1.1e-06 & \textbf{7.6e-07} & 7.7e-07 & \\
Llama3.1-8B           & .0072  & .5098 & - & \textbf{.7286} & 4.9e-06 & \textbf{1.9e-06} & 3.6e-06\\
% Llama2-13B            & -.1973 & .7261 & .1141 & \textbf{.9965} & 2.1e-06 & 1.5e-06 & 4.3e-04 & \textbf{6.0e-08}\\
Llama2-70B            & .0283  & n/a\protect\footnotemark
& n/a & \textbf{1.000} & 2.2e-04  & n/a & \textbf{5.9e-07}\\
\midrule
Avg. Runtime (Llama2-7B) & \textbf{0.149s}  & 19.349s & 6.086s & 0.161s & \multicolumn{4}{c}{Same as left}\\
\bottomrule
\end{tabular}
\caption{Evaluation of NeurGrad and baselines in calculating NEGs, including two metrics: $r$ and MAE. }
\label{tab:grad_compare}
\end{table*}
\addtocounter{footnote}{-1}
\footnotetext{We follow code released in \citet{yu-ananiadou-2024-neuron} and only Llama2 LLMs are supported.}
\addtocounter{footnote}{1}
\footnotetext{Due to the high memory cost of IG and LPI, we preclude Llama2-70B experiments on these methods.}
% \subsection{Evaluation of NeurGrad}

\subsection{\shin{NEG Estimation Evaluation}}
We evaluate NeurGrad's ability to estimate NEG using the same setup as in the CG evaluation.
% above to collect the ground-truth NEGs.}
% \shin{The ground-truth NEGs are collected from the intervention experiments on 1000 prompts, each with 100 random neurons, and the activation shift range is set to $\pm$2.
% First, we obtain ground-truth NEG through neuron-wise intervention experiments.
% Then, we measure the Pearson correlation between ground truth and NeurGrad-estimated NEG. 
% Specifically, we collect empirical gradients of 1000 prompts, with 100 random neurons per prompt.
% The activation shift range is set to [-2, 2] according to \S~\ref{neuron-linearity}.
\shin{Our experiment compares NeurGrad with three baselines: two gradient-based methods, CG and integrated gradients (IG)~\cite{pmlr-v70-sundararajan17a,dai-etal-2022-knowledge}, and one logit-based method (LPI)~\cite{yu-ananiadou-2024-neuron}.
IG simulates NEG by small, repeated neuron interventions, while LPI (Log-Probability-Increase) estimates neuron importance based on increases in output probabilities.\footnote{We exclude causal-tracing methods~\citep{meng2022locating}, which are too costly and do not meet our efficiency goals.}}
% \shin{We also conduct the evaluation for integrated gradients (IG) used for identifying knowledge neurons~\citep{dai-etal-2022-knowledge} that intervene neurons in small step sizes multiple times to simulate the gradient.}
% \footnote{Note that while studies like \citet{meng2022locating,geva-etal-2022-transformer,yu-ananiadou-2024-neuron} can also assign attribution scores to neurons in impacting the output, they are not directly for estimating the gradients, so we don't compare them due to the different objectives.}.}

\begin{figure}[t]
    \centering
    \includegraphics[width=\linewidth]{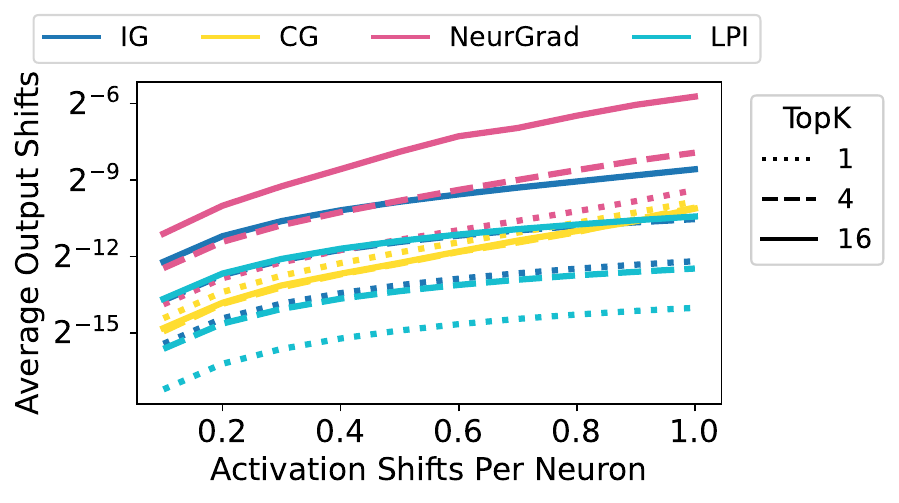}
    \caption{Comparison of neuron attribution methods in token probability enhancement. \textit{X-axis}: activation shifts of selected neurons; \textit{Y-axis}: average output shifts over 1000 factual prompts.}
    \label{fig:attribution_method_comparision}
\end{figure}

\shin{We assess NEG estimation using correlation ($r$) and mean absolute error (MAE) against ground-truth NEG\@. 
% The prior assesses NeurGrad's ability to capture the relative relationship between neurons, like ordering, while MAE directly quantifies its accuracy of NEGs' calculation.
As show in Table~\ref{tab:grad_compare},
%(left) shows the Corr between the estimated and ground-truth NEGs that can assess their ability to capture the neurons' relative relationship, such as ordering. 
NeurGrad consistently achieves the highest $r$ across the seven PLMs, capturing relative neuron importance well.
% outperforming all other methods. 
% Moreover, Table~\ref{tab:grad_compare} (right) reports the Corr the MAE that quantifies the accuracy of NEGs' calculation. It is observed that 
NeurGrad also yields low MAE,
% can largely reduce the estimation error to the true NEG, revealing its ability to precisely calculate NEGs.
indicating high accuracy.
The average running time of Llama2-7B with an NVIDIA RTX A600 GPU (Table~\ref{tab:grad_compare} (bottom)) demonstrates NeurGrad's superior efficiency.}
% The time estimation uses Llama2-7B with an NVIDIA RTX A6000 GPU.}

% Regarding efficiency, calculating IG requires multiple iterations, while NeurGrad completes the calculation with just one inference pass, resulting in a computational cost nearly identical to that of computational gradients.
% \shin{We also evaluate the effectiveness of using NeurGrad to find important neurons, see \S~\ref{sec:appendix-knowledge-attribution-evaluation} for details.}

\subsection{\shin{Knowledge Attribution Evaluation}}
\label{sec:knowledge-attribution-evaluation}

\shin{We further evaluate NeurGrad's ability to identify important neurons. 
For 1000 prompts, we select the top-$K$ neurons ($K=1, 2^2, 2^4$) using CG, IG, LPI, and NeurGrad values, and \cmr{then enhance their activations by increasing activations of positive neurons and decreasing those of negative neurons. The activation shift is conducted within the range of [0.1, 1] with a step size of 0.1.
Figure~\ref{fig:attribution_method_comparision} shows output shifts on Llama2-7B under these interventions.}
NeurGrad consistently outperforms baselines, due to its accurate NEG estimation and inclusion of both positive and negative neurons, unlike IG and LPI, which only consider positive ones, despite their equal distribution (Table~\ref{tab:neuron_ratios}). See \S~\ref{sec:appendix-supplementary-knowledge-attribution} for details.}

\begin{figure}[t]
    \centering
    \includegraphics[width=\linewidth]{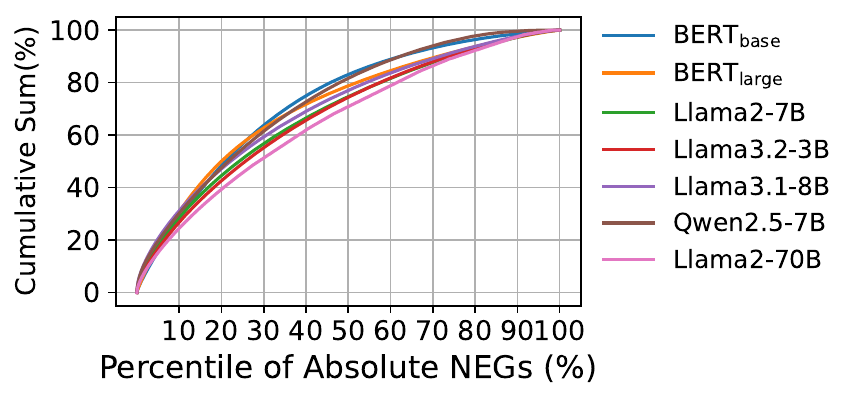}
    \caption{Cumulative distribution of NEG magnitudes.
    (\textit{X-axis}: the percentiles of NEG magnitudes; \textit{Y-axis}: the cumulative contribution of neurons to the total sum).}
    \label{fig:grad_percentile}
\end{figure}

\section{Understanding Neurons' Controllability}
\label{sec:understand-neuron-controllability}

This section explores neuron controllability: the ability to precisely adjust PLM output probabilities by modifying neuron activations. We use NEGs estimated by NeurGrad to achieve this.

% \begin{table}[t]
%     \centering
%     \small
%     \begin{tabular}{lcc}
%         \toprule
%         & \textbf{BERT$_\mathrm{base/large/wwm}$} & \textbf{Llama2-7/13/70B} \\
%         \midrule
%         Pos. ratio & .5019/.5008/.4996 & .4604/.4664/.4484 \\
%         Neg. ratio & .4981/.4992/.5004  & .4592/.4660/.4480 \\ 
%         \bottomrule
%     \end{tabular}
%     \caption{The pos/neg neuron ratios over 1000 prompts.}
%     \label{tab:pos_neg_ratio}
% \end{table}

% \subsection{Distribution of Neurons' NEG}
\subsection{How Are NEGs Distributed?}
\label{sec:neuron-distribution}
\paragraph{Do only a few neurons exhibit strong gradients?}
Figure~\ref{fig:grad_percentile} presents the cumulative NEG distribution for all neurons, showing
% Rising curves are 
steady growth that only converges when all neurons are included. This indicates that most neurons influence the model's output probabilities. 

\paragraph{Do neurons have polarity preference?}
\label{sec:neg-distributions}
Table~\ref{tab:neuron_ratios} (\S~\ref{sec:neuron_analysis}) shows the ratios of positive and negative neurons\footnote{We used ground-truth NEGs here to avoid flipping the polarity of neurons with NEGs near zero.} across 1000 prompts, each with 100 randomly sampled neurons in seven PLMs, revealing nearly equal numbers.
% that the numbers of positive/negative neurons are nearly equivalent. 
This suggests that PLMs have no polarity preference; interventions should consider gradient polarity rather than simply 
% enhancing or suppressing neurons should be guided by their gradient polarity rather than merely 
increasing or decreasing their activations~\cite{dai-etal-2022-knowledge}.
See more detailed analysis on NEGs' distribution in \S~\ref{sec:appendix-gradient-distribution} and \S~\ref{sec:appendix-neuron-layer-distribution}.

% It demonstrates that not a limited number of neurons but a large majority of neurons have the ability to adjust output probabilities.
% Notably, Llama2 models exhibit smoother and more evenly distributed curves than BERTs, implying that neuron-level controllability is more widely dispersed in Llama2.}

\begin{figure}[t]
    \centering
    \includegraphics[width=\linewidth]{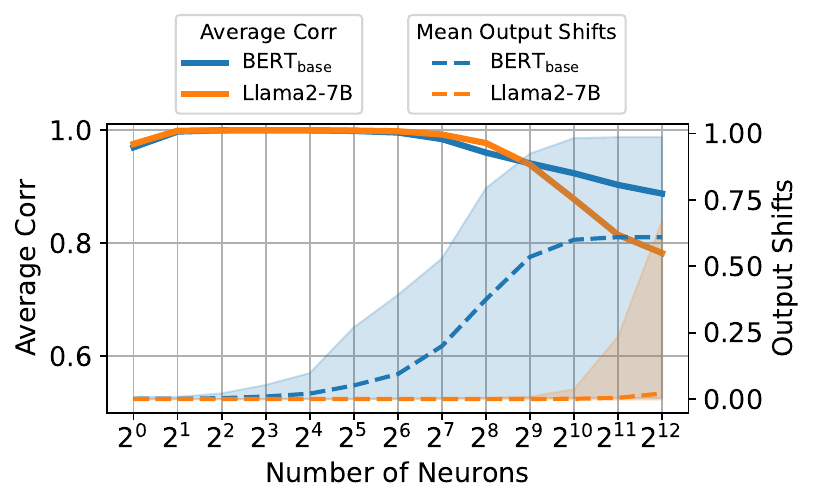}
    \caption{Multi-neuron enhancement with range [0,0.5] with different number of neurons.}
    % \textbf{Y-axis (left)}: average Corr between activation shifts per neuron and total output shifts; \textbf{Y-axis (right)}: the range and average of output shifts.}
    \label{fig:multi-neuron}
\end{figure}

\subsection{Does Linearity Hold for Multi-neuron?}
\label{sec:multineuron-intervention}
% NeurGrad's ability to accurately measure NEG raises the question: Can output shifts be predicted when multiple neurons are shifted?
We explore 
\textbf{whether output shifts can be predicted when intervening multiple neurons.} For each of 1000 prompts, 
we randomly sample $N$ neurons and shift
% enhance them by shifting 
their activations % according to 
on the basis of polarity measured by NeurGrad, applying
% in which 
positive/negative shifts to positive/negative neurons.
We experiment on BERT$_\mathrm{base}$ and Llama2-7B using neuron sizes of $2^N$ ($0 \leq N \leq 12$).
% The enhancement range is set as [0, 0.5] with a step size of 0.01.

\shin{Figure~\ref{fig:multi-neuron} shows the average correlation ($r$) between predicted and actual output shifts across all prompt-neuron pairs, using an enhancement range of [0, 0.5] and a 0.01 step size. 
While $r$ decreases as more neurons are involved due to neuron interactions, 
% The weakening of linearity is likely due to neuron interactions across different Transformer layers. 
it remains strong ($\geq$0.7) even with $2^{12}$ neurons in both PLMs. 
The figure also
% Figure~\ref{fig:multi-neuron} (right y-axis)
shows that larger neuron sets
% involving more neurons can 
cause greater output shifts, suggesting an additive effect.
% that the output shifts caused by individual neurons can be accumulated. 
% Note that despite 
Although Llama2 shows small average output shifts (due to the lower per-neuron NEG; see \S~\ref{sec:appendix-gradient-distribution}), significant changes still occur when over $2^{10}$ neurons are intervened.
% involved in Llama2-7B models with specific neuron combinations.
% output shift range increases rapidly for BERT, but not for Llama2-7B, due to smaller NEG magnitudes in Llama2 models (see \S~\ref{} for details). 
% Moreover, our analysis 
Experiments with larger enhancement ranges show similar trends: greater shifts load to lower $r$, reducing predictability (see \S~\ref{sec:appendix-supplementary-multineuron-intervention} for experiments with different ranges).}

\shin{The analysis above shows that NeurGrad enables partial prediction of output shifts. However, both the number of modified neurons and the modification range require careful consideration.}

\subsection{\cmr{Why Do Neurons Exhibit Linearity?}}
\label{sec:appendix-local-linearity}
We propose the local linearity approximation hypothesis to explain neuron linearity, based on three observations: i) larger shift ranges reduce $r$, ii) larger PLMs show higher $r$ due to the reduced influence of individual neurons (Figure~\ref{fig:shift_range}), and iii) involving more neurons weakens linearity (Figure~\ref{fig:multi-neuron}).
This aligns with the first-order Taylor expansion under local differentiability.

\begin{tcolorbox}[
fontupper=\ttfamily, 
title={Local Linear Approximation Hypothesis},
    boxsep=1pt,
    left=4pt, right=4pt, top=4pt, bottom=4pt,
    colback=white, 
    colframe=black,
    before skip=5pt, after skip=5pt]
%v0
% Let $f: \mathbb{R}^n \to \mathbb{R}^m$ be a PLM, where $x \in \mathbb{R}^n$ denotes neuron activations and $f(\bm{x})$ the output token probabilities. 
% The effect of neuron $x_i$ in a local region can be linearly approximated as:
% $$
% f(x_i + \delta e) \approx f(x_i) + \frac{\partial f}{\partial x_i} \delta,
% $$
% where $e_i$ is the unit vector. 

%v1
% Let $\mathbf{f}: \mathbb{R}^n \rightarrow \mathbb{R}^m$ denote the mapping from a model’s FF activation layer $\mathbf{x} \in \mathbb{R}^n$ to its output probability vector $\mathbf{f}(\mathbf{x})$. For a perturbation on $i$-th neuron $x_i$, define
% $$
% \mathbf{x}' = \mathbf{x} + \Delta x_i \mathbf{e}_i,
% $$
% where $\mathbf{e}_i$ is a standard basis vector with a 1 in position $i$ and zeros elsewhere, and $\Delta x_i \in \mathbb{R}$ is the finite change.

%v2
Let $\mathbf{f}: \mathbb{R}^n \rightarrow \mathbb{R}^m$ denote the mapping from a model’s FF activation layer $\mathbf{x} \in \mathbb{R}^n$ to its output probability vector $\mathbf{f}(\mathbf{x})$. We have:
\begin{align*}
f_j(\mathbf{x}') 
&= f_j(\mathbf{x}) + \frac{\partial f_j}{\partial x_i}(\mathbf{x})\, \Delta x_i + o(\Delta x_i)
\end{align*}
if we consider a perturbation $\Delta x_i$ on $i$-th neuron, with perturbed vector $\mathbf{x}' = \mathbf{x} + \Delta x_i \mathbf{e}_i$. Here $(\mathbf{e}_i)_j=\delta_{ij}$ is a unit vector. Then, we define $\frac{\partial f_j}{\partial x_i}(\mathbf{x})$ as \textbf{NEG}.
\end{tcolorbox}

\section{Skill Neuron Probing using NeurGard}
\label{sec:skill-neuron-probing-task}
% Neurons can linearly influence token probabilities, providing a reliable way to control PLM outputs. 
% This raises the question of whether neurons can act as hubs for aligning model output with desired knowledge representations.
% empirical gradients can effectively represent language knowledge. 
% We investigate this by conducting experiments to localize skill neurons based on their empirical gradients. 
% We have demonstrated that neurons can exert a linear influence on output probabilities in factual probing tasks, highlighting their potential to manipulate model output probabilities predictably.
\shin{While NEG and NeurGrad allow neuron-level output control, variation in NEG across prompts limits their use for modifying specific language skills, which require
%. These skills often involve 
handling diverse inputs. 
This section investigates \textbf{whether NEG can capture general language skills through skill neuron probing,}}
% Building on this, we investigate whether NEG can effectively encode diverse language skills through the skill neuron probing~\citep{wang-etal-2022-finding-skill}. 
which identifies neurons linked to task-solving capabilities.
% that encode the ability to solve language tasks. 
Unlike prior work~\citep{wang-etal-2022-finding-skill,song-etal-2024-large} focusing on neuron activations, 
% explored the effectiveness of using neuron activations, yet 
we utilize NEG for finding skill neurons\@.
% had different research objectives from ours.

\subsection{Task Definition}

Following \citet{wang-etal-2022-finding-skill}, we define skill neuron probing as follows. 
A skill dataset 
% conveying specific language skills 
$\mathcal{D}$ consists of language sequence pairs: knowledge inquiries $\mathcal{Q}=\{q_1, ..., q_{|\mathcal{T}|}\}$ and corresponding answers $\mathcal{A}=\{a_1, ..., a_{|\mathcal{T}|}\}$, where 
each $a_i$ is from a candidate set $\mathcal{\hat{A}}_{\text{cands}}$.
For example, in sentiment classification, $Q$ is the set of documents and $A$ the sentiment labels. 
We train classifiers using the behavior of a  neuron subset $\mathcal{N}_s \subseteq \mathcal{N}$
 as features to predict the correct answer $a_i$ for each $q_i$, where 
$\mathcal{N}$ is the full neuron set.
\footnote{We focus on intermediate outputs (neurons) of FF layers.} 

Our skill neuron probe finds the optimal neuron subset $\mathcal{N}_s^*$ that maximizes accuracy on the dataset $\mathcal{D}$.
\begin{equation}
\mathcal{N}_s^* = \underset{\mathcal{N}_s \subseteq \mathcal{N}}{\arg\max} \, \text{Acc}(f(\mathcal{N}_s), D)
\end{equation}
\begin{equation}
\quad \text{Acc}(f(\mathcal{N}_s), D) = \frac{1}{|D|} \sum_{i=1}^{|D|} \mathbbm{1}[f(\mathcal{N}_s, q_i) = a_i].
\end{equation}
Here, $f(\mathcal{N}_s, q_i)$ is the classifer's prediction using neuron subset $\mathcal{N}_s$ for input $q_i$ and
$\mathbbm{1}[X = Y]$ is $1$ if $X=Y$, otherise $0$.

\subsection{Evaluation Benchmark: MCEval8K}
\label{sec:mceval8k}
In this section, we introduce a benchmark dataset for skill neuron probing.
Since probing requires a fixed target token, previous work~\citep{wang-etal-2022-finding-skill,song-etal-2024-large} used on multiple-choice datasets with single-token labels (\textit{e.g.}, A: positive, B: negative). 
While we follow a similar setup, earlier studies focused on small PLMs and probed them on basic tasks, which are insufficient for evaluating LLMs.

To address this limitation, we create MCEval8K, a diverse multiple-choice benchmark covering 22 language understanding tasks across six skill genres, incorporating most datasets from previous studies.
% difficulties in matching answer sequences and candidate category labels in classification tasks.
% To mitigate this,
% we restrict the $a_i$ to a single token and limit option numbers.
% To meet these requirements, 
% We thus create a multi-choice language skill evaluation benchmark, MCEval8K, that forces PLMs to generate a single-token option label (A, B, etc.) named for category labels (A: positive, B: negative, etc.). 
% MCEval8K encompasses 22 tasks across 6 distinct genres, conveying diverse language skills to evaluate the neurons' representation capacity.
% For tasks that are not multi-choice datasets, we construct several options for each knowledge inquiry. 
To ensure consistency and reduce computational cost, we cap each task at 
%Since tasks vary in different sizes, with some, such as cLang-8~\citep{clang8-1, clang8-2}, containing millions of data points, we standardize the evaluation by limiting each task to 
8K queries,\footnote{Only the Stereoset task has fewer than 8K queries due to the limited size of the original dataset.}
even for large datasets like cLang-8~\citep{clang8-2,clang8-1}.
% It minimizes unnecessary computational costs while ensuring consistency across tasks.
We also balance the number of correct options per task to avoid classification bias.
% introduced by imbalanced classification.
Task and dataset details 
% The skill genres, tasks, and datasets information 
are provided below and in \S~\ref{sec:appendix-mceval8k}. 
% for more dataset creation details.

\smallskip\noindent\textbf{Linguistic:} 
Part-of-speech tagging (POS), text chunking (CHUNK), named entity recognition (NER), and grammatical error detection (GED). 

\smallskip\noindent\textbf{Content classification:} 
Sentiment (IMDB), topic (Agnews), and Amazon reviews with numerical labels (Amazon).

\smallskip\noindent\textbf{Natural language inference (NLI):} Textual entailment (MNLI), paraphrase identification (PAWS), and grounded commonsense inference (SWAG). 

\smallskip\noindent\textbf{Factuality:}
Fact-checking (FEVER), knowledge probing (MyriadLAMA), commonsense QA (CSQA), and temporal fact probing (TempLAMA).

\smallskip\noindent\textbf{Self-reflection:} %Examine PLMs' internal status, including 
Hallucination (HaluEval), toxicity (Toxic), and stereotype detections (Stereoset).

\smallskip\noindent\textbf{Multilinguality:} 
Language identification (LTI), multilingual POS tagging (M-POS), sentiment classification (M-Amazon), knowledge probing (mLAMA), and textual entailment (XNLI).

\begin{figure*}[t]
    \centering
    \includegraphics[width=0.9\linewidth]{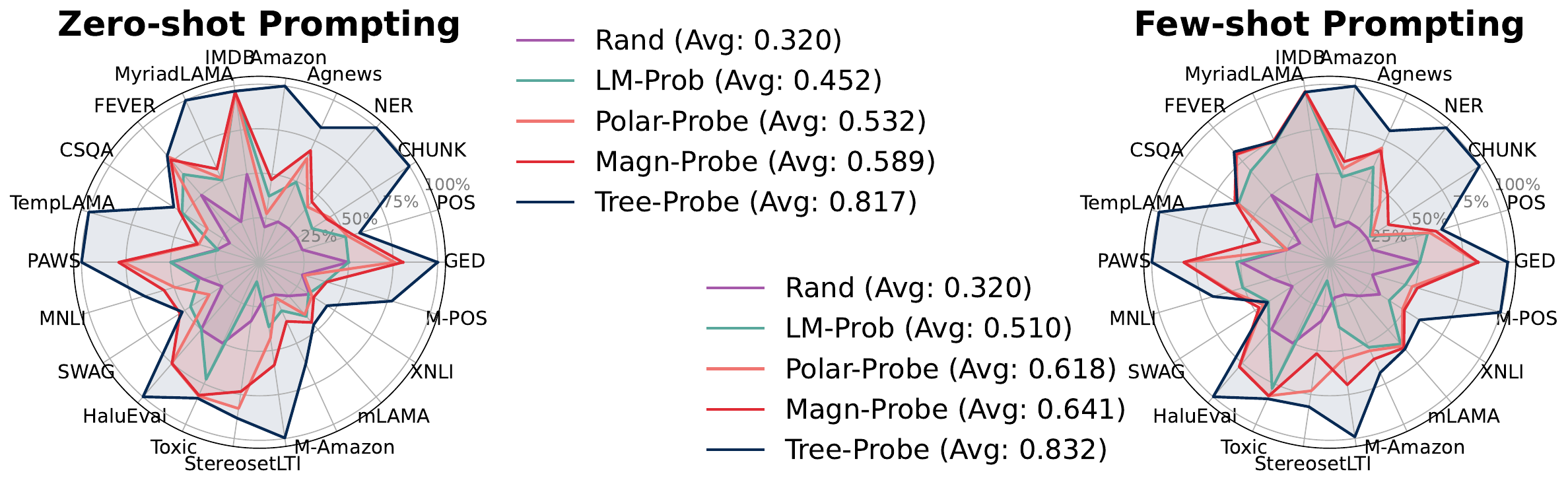}
    \caption{MCEval8K accuracies on Llama2-7B across tasks in zero-shot and few-shot settings, reported for Rand (random guess), LM-Prob (token probability), and three proposed probes. Legends show average accuracies. }
    \label{fig:full_acc_rader}
\end{figure*}

\begin{table*}[t]
    \centering
    \small
    \setlength{\tabcolsep}{3.1pt}
    \begin{tabular}{lcccccccccccccccccc}
        \toprule
        \textbf{Model} &
        \multicolumn{3}{c}{\textbf{NER}} &
        \multicolumn{3}{c}{\textbf{Agnews}} &
        \multicolumn{3}{c}{\textbf{PAWS}} &
        \multicolumn{3}{c}{\textbf{CSQA}} &
        \multicolumn{3}{c}{\textbf{HaluEval}} &
        \multicolumn{3}{c}{\textbf{mLAMA}} \\
        \cmidrule(lr){2-4}
        \cmidrule(lr){5-7}
        \cmidrule(lr){8-10}
        \cmidrule(lr){11-13}
        \cmidrule(lr){14-16}
        \cmidrule(lr){17-19}
        & LM & Act & Mag & LM & Act & Mag & LM & Act & Mag & LM & Act & Mag & LM & Act & Mag & LM & Act & Mag \\
        \midrule

        \textbf{Llama3.2-3B} & 
        .598 & \textbf{.633} & .618 & 
        .757 & \textbf{.857} & .776 & 
        .500 & \textbf{.843} & .831 & 
        .551 & .681 & \textbf{.691} & 
        .650 & \textbf{.822} & .780 & 
        .473 & .563 & \textbf{.570} \\

        \textbf{Llama2-7B} & 
        .361 & .453 & \textbf{.498} & 
        .588 & \textbf{.849} & .702 & 
        .524 & \textbf{.825} & .815 & 
        .610 & .613 & \textbf{.639} & 
        .520 & \textbf{.788} & .783 & 
        .608 & .622 & \textbf{.637} \\

        \textbf{Qwen2.5-7B} & 
        .871 & \textbf{.877} & \textbf{.877} & 
        .751 & \textbf{.858} & .755 & 
        .574 & \textbf{.889} & .873 & 
        .768 & \textbf{.816} & .814 & 
        .637 & \textbf{.821} & .773 & 
        .784 & \textbf{.795} & .785 \\
        
        \textbf{Llama3.1-8B} & 
        .815 & .826 & \textbf{.833} & 
        .770 & \textbf{.883} & .812 & 
        .500 & \textbf{.860} & .854 & 
        .674 & .733 & \textbf{.740} & 
        .674 & \textbf{.807} & .798 & 
        .762 & .781 & \textbf{.785} \\
        
        \textbf{Llama2-70B} & 
        .790 & - & \textbf{.817} & 
        .763 & - & \textbf{.824} & 
        .779 & - & \textbf{.846} & 
        .754 & - & \textbf{.763} & 
        .753 & - & \textbf{.825} & 
        .743 & - & \textbf{.760} \\
        \bottomrule
    \end{tabular}
    \caption{Accuracy of six tasks across five LLMs: \textbf{LM}-Prob, \textbf{Act}ivation-based, and \textbf{Mag}n-Probe.}
    \label{tab:model-compare-acc}
\end{table*}

%\subsection{NEG as Knowledge Feature}
\subsection{NEG-based Skill Neuron Probe}
\label{sec:skill-neuron-experiment}
We train skill neuron probes using NeurGrad's estimated NEG to examine how NEG encodes general language skills.
%\subsubsection{NEG-based Skill Neuron Probe}
Each skill dataset $\mathcal{D}$ is split into: training, validation, and test sets ($\mathcal{D}_\mathrm{train}$, $\mathcal{D}_\mathrm{valid}$, and $\mathcal{D}_\mathrm{test}$), with a ratio of 6:1:1. 
We compare the following three probes with different designs. 

\smallskip\noindent\textbf{Polar-Probe} is a majority-vote classifier, where each neuron votes based on its polarity (positive or negative).
For each training pair $(q_i, a_i)$ in $\mathcal{D}_{\mathrm{train}}$ and neuron $n_k$, we record its polarity as feature $\bm{x}_{q_i, a_i}^{n_k}$.
%For each $n_k$, we calculate the ratio of being positive and negative 
The dominant polarity across $\mathcal{D}_\mathrm{train}$ 
%examples and the dominant polarity is identified as their 
defines the neuron's global polarity $\bar{\bm{x}}^{n_k}$, and neurons with more consistent polarity are ranked higher. 
% For instance, if $n_j$ exhibits positive polarity in more than half of the $(q_i, a_i)$ pairs, its global polarity is classified as positive.

% After identifying the global polarity and neuron ranks, we use $\mathcal{D}_{\text{valid}}$ to determine the skill neurons $\mathcal{N}_s^*$. 

To predict $q_i$, we compare polarities $\bm{x}_{q_i, a_j}^{n_k}$ for each candidate $a_j \in \hat{\mathcal{A}}_{\text{cands}}$ and $n_j \in \mathcal{N}_s^*$:
% The prediction is:
\begin{align}
f(\mathcal{N}_s^*, q_i)&= \underset{a_j \in \hat{\mathcal{A}}_{\text{cands}}}{\arg\max} \sum_{n_k \in \mathcal{N}_s^*} \mathbbm{1}[\bm{x}_{q_i, a_j}^{n_k}=\bar{\bm{x}}^{n_k}]\label{eq:major-vote}
\end{align}
The optimal size of $\mathcal{N}_s^*$ is selected using $\mathcal{D}_{\text{valid}}$.

\smallskip\noindent\textbf{Magn-Probe} 
\label{para:magnitude-probe} uses NEG magnitudes as features for a majority-vote classifier.
For each training pair $(q_i, a_i)$ in $\mathcal{D}_\mathrm{train}$ and neuron $n_k$, we compare NEGs across $a\in \hat{\mathcal{A}}_{\text{cands}}$. 
Neurons that consistently show the highest or lowest NEGs for the correct answer $a_i$ are marked as skill indicators, along with their preference for being highest or lowest. 
More consistent neurons are ranked higher. 
At inference, predictions follow the same voting rule as in Eq.~\ref{eq:major-vote}.
This probe evaluates whether NEG magnitudes can encode skill information. 
% aiming to investigate the differences between using polarity and gradient magnitude as feature sources.

\smallskip\noindent\textbf{Tree-Probe} 
\label{para:dtree-probe} 
\cmr{is designed to analyze the impact of accounting for interdependencies among skill neurons.
We use the index (non-negative integers) of $a \in \hat{\mathcal{A}}_{\text{cands}}$ with the most significant NEGs as features for training a random forest classifier. 
The hyperparameters include the number of trees (\texttt{\#n\_trees}) and layers (\texttt{\#n\_layers}) used in each tree.
See more details in \S~\ref{sec:appendix-random-forest-probe}}.

\subsection{Experimental Setup}
\label{subsec:probing_experiment_setp}

\smallskip\noindent\textbf{Dataset \& Prompt.}
\cmr{We split each MCEval8K dataset into train, validation, and test by the ratio of 6:1:1, ensuring balanced correct-answer tokens across subsets to avoid majority label bias~\cite{zhao2021calibrateuseimprovingfewshot}.
We manually craft instructions and options for all MCEval8K tasks to ensure single-token outputs, considering both zero-shot and few-shot settings.}
See \S~\ref{sec:appendix-prompt-instructions} for all task instructions.

\paragraph{Probe.}
For majority-based probes, we select the optimal neuron size from $2^n (0\leq n\leq 13)$ using $\mathcal{D}_\mathrm{valid}$.
\cmr{For the Tree-Probe (random forest), we report accuracy using scikit-learn's default settings, where the optimal subset of features is selected automatically: 100 trees with no depth limitation.}
See \S~\ref{sec:appendix-per-task-result},~\ref{sec:appendix-random-forest-probe} for details.

% \begin{table}[t]
%     \centering
%     \small
%     \setlength{\tabcolsep}{5.5pt}
%     \begin{tabular}{lcc}
%     \toprule
%     Methods & \textbf{Zero-shot} & \textbf{Few-shot} \\ \midrule
%     \textbf{Random Guess (RG)}       & \multicolumn{2}{c}
%     {.3302} \\
%     \textbf{Token Probability (TP)} &  .4616 & .5163 \\
%     \textbf{Polar-Probe}         &  .5559 & .6331 \\
%     \textbf{Magn-Probe}          &  .6096 & .6495 \\
%     \textbf{Tree-Probe}          &  .8286 & .8345 \\
%     \bottomrule
%     \end{tabular}
%     \caption{Average MCEval8K accuracy on Llama2-7B.}
%     \label{tab:average-acc}
% \end{table}

\paragraph{Models.}
\cmr{We probe Llama2-7B with three NEG-based probes on all tasks in MCEval8K. 
For other LLMs, to reduce computational cost, we only probe one dataset per genre (NER, Agnews, PAWS, CSQA, HaluEval, and mLAMA) with $2^{10}$ training examples, focusing on majority-vote probes.}

\subsection{Result and Analysis}
\cmr{We compare our skill probes with three baselines: \textbf{Rand} (random guessing, \textbf{LM-Prob} (choosing the token with highest model probability), and \textbf{Act} (activation-based probes from \citet{wang-etal-2022-finding-skill}).}

\paragraph{Can NEG capture language skills?} 
Figure~\ref{fig:full_acc_rader} shows Llama2-7B’s accuracy on MCEval8K tasks in zero- and few-shot settings. 
\cmr{The results show that LM-Prob consistently outperforms random guessing, indicating that Llama2-7B can follow instructions to activate the relevant language skills. 
However, even the simple majority-vote probes can surpass the LM-Prob, suggesting that NEGs capture meaningful information about language skills in individual neurons.
Furthermore, Tree-Probe significantly outperforms majority-vote probes by integrating NEGs from multiple neurons. This highlights that the interplay among neurons is crucial for accurately representing language skills.
See Tables~\ref{tab:per-task-acc-zeroshot} and \ref{tab:per-task-acc-fewshot} in Appendix \S~\ref{sec:appendix-per-task-result} for details on all tasks.
Experiments on other LLMs further validate our findings as reported in Table~\ref{tab:model-compare-acc}.}

% \begin{table}[t]
%     \centering
%     \small
%     \setlength{\tabcolsep}{2pt}{
%         \begin{tabular}{lcccc}
%         \toprule
%         \multirow{2}{*}[-2pt]{\textbf{Tasks}} & 
%         \multicolumn{2}{c}{\textbf{Llama2-7B}} & 
%         \multicolumn{2}{c}{\textbf{Llama2-70B}} \\
%         \cmidrule(lr){2-3} \cmidrule(lr){4-5}
%         &LM-Prob & Magn-Probe & LM-Prob & Magn-Probe \\ \midrule
%         \textbf{NER} & .3610 & .4980 & .7900 & \textbf{.8170} \\
%         \textbf{Agnews} & .5880 & .7020 & .7630 & \textbf{.8240} \\
%         \textbf{PAWS} & .5240 & .8150 & .7790 & \textbf{.8460} \\
%         \textbf{CSQA} & .6100 & .6390 & .7540 & \textbf{.7630} \\
%         \textbf{HaluEval} & .5200 & .7830 & .7530 & \textbf{.8250} \\
%         \textbf{mLAMA} & .6080 & .6370 & .7430 & \textbf{.7600} \\
%         \bottomrule
%         \end{tabular}
%     }
%     \caption{Accuracy of six tasks on Llama2-7B and -70B\@.}
%     \label{tab:model-compare-acc}
% \end{table}

% \smallskip\noindent\textbf{Skill neurons exhibit interdependency.} 
% The major-vote probe assumes independence between neurons, while the Tree-Probe models their interdependencies and builds hierarchical classifiers. 
% Its significant advantage over the major-vote probe in Table~\ref{tab:average-acc} suggests that all neurons do not independently represent language skills, but interplay exists between them (See \S~\ref{sec:appendix-neuron-hierarchy} for a deeper analysis).

\paragraph{Which encodes skills better: activation or gradient?} 
\cmr{Table~\ref{tab:model-compare-acc} demonstrates that NEG-based probes have a different focus from the activation-based approach. Specifically, the gradient-based probes perform better on knowledge-intensive language tasks, such as factual knowledge and commonsense reasoning (mLAMA and CSQA). 
One possible explanation is that simple linguistic knowledge becomes saturated during pretraining, while complex knowledge remains undertrained, making it more effectively captured by NEG-based probes.}

\cmr{Furthermore, NEG-based probes outperform activation-based methods in efficiency by assigning distinct values to different target tokens per neuron, enhancing representational capacity. In contrast, activation-based approaches rely only on forward pass signals, limiting neurons to binary distinctions and making them less effective for multi-class tasks~\cite{wang-etal-2022-finding-skill}.}

% \paragraph{Larger PLMs better activate skills.} 
% Table~\ref{tab:model-compare-acc} compares LM-Prob and Magn-Probe on six few-shot tasks for Llama2-7B and -70B. While LLaMA2-70B outperforms 7B under LM-Prob, the gap nearly disappears with Magn-Probe. This suggests that 70B’s superior performance comes not just from storing more knowledge, but from more effectively activating and expressing it in outputs.

% Skill neuron probe also shows closer performance between the two models compared to prompting, suggesting that small and large models store knowledge more similarly than prompting reveals. 

% \paragraph{Interpreting in-context learning with empirical gradients.}
\paragraph{How do NEGs make predictions?}
We analyze NEGs associated with each option token to explore why simple majority-vote classifiers can achieve high accuracy. 
Using PAWS (binary classification) as an example, we acquire NEGs for target tokens (``\textit{yes}/\textit{no}'') across 6K prompts. 
We find 97.21\% of neurons show opposite signs for ``\textit{yes}/\textit{no}'' tokens, with a near-perfect negative correlation ($-0.9996$), indicating strong polarity across options.
\shin{We further compare the NEGs given zero-shot and few-shot prompts. 
We find that NEG magnitudes of few-shot prompts over 22 tasks are 5.36 times larger than zero-shot, suggesting that adding demonstrations can effectively activate skill neurons.}

\section{Properties of Skill Neurons}
\label{sec:skill-neuron-property}

To deepen our understanding of how language skills are represented within neurons, we analyze skill neurons' behavioral and structural properties through a series of focused studies.

% The previous section shows that empirical gradients of specific neuron groups represent language skills. 
% This section dives deep into the properties of skill neurons, investigating the distribution and interaction of skill neurons.
% within LLMs remain unexplored. 
% This section investigates these properties further.

\begin{table}[t]
    \centering
    \footnotesize
    \setlength{\tabcolsep}{4.5pt}
    \begin{tabular}{ll}
    \toprule
    \textbf{\makecell[l]{Neuron sizes}} & \textbf{Tasks} \\ \midrule
    \textbf{$2^0 \sim 2^3$}      & Toxic, LTI, M-POS, FEVER, TempLAMA\\
    \midrule
    \textbf{$2^4 \sim 2^8$}      & \makecell[l]{GED, POS, CHUNK, NER, \\Amazon, IMDB, PAWS, MNLI, \\SWAG, HaluEval, XNLI, M-Amazon} \\
    \midrule
    \textbf{$2^9 \sim 2^{13}$}   & Agnews, MyriadLAMA, CSQA, mLAMA \\
    \bottomrule
    \end{tabular}
    \caption{The optimal neuron sizes for Mag-Probe.}
    \label{tab:optimal-neurons}
\end{table}

\begin{figure}[t]
    \centering
    \includegraphics[width=\linewidth]{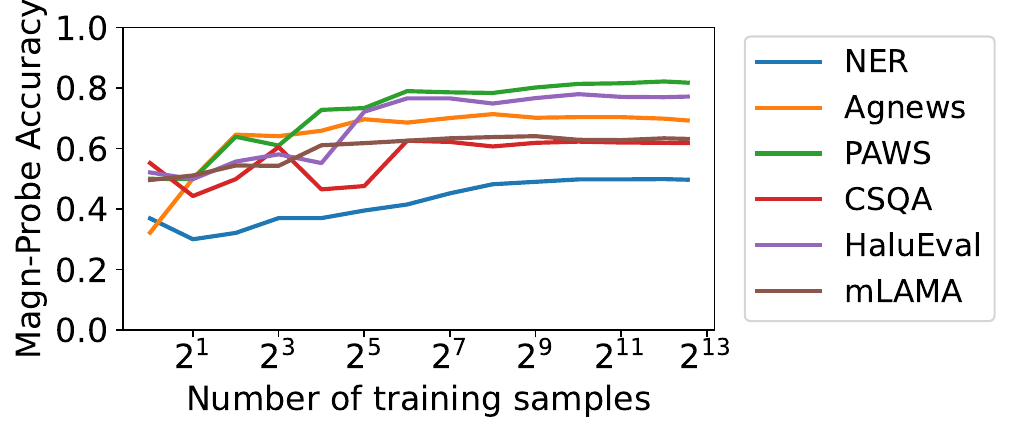}
    \caption{Accuracies with varying training sizes.}
    \label{fig:trainsize2accuracy}
\end{figure}

\subsection{Representation \& Acquisition Efficiency}
\label{sec:neuron-efficiency}

\paragraph{Representational efficiency:} \cmr{By tuning neuron size on the validation set, we find that skill-neuron probes achieve high accuracy with only a few neurons.
The optimal neuron sizes for all tasks with Magn-Probe are reported in Table~\ref{tab:optimal-neurons}. 
Most tasks can achieve optimal accuracy within 256 neurons, demonstrating the efficiency of NEG in representing language skills. 
Notably, factuality tasks, such as MyriadLAMA, CSQA, and mLAMA, engage more neurons, suggesting that handling facts requires more diverse neurons, reflecting the complexity of factual understanding tasks.}

\paragraph{Acquisition efficiency:} 
\cmr{Figure~\ref{fig:trainsize2accuracy} reports the accuracy of skill-neuron probes with different training examples. 
While adding training examples can consistently increase the probes' accuracy, the earnings slow down after 128, indicating the efficiency of acquiring skill neurons with limited data. }

\subsection{How Robust Are Skill Neurons to Contextual Variations?}
\label{sec:neuron-generality}
\cmr{We investigate how skill neurons change when we provide different contexts, including instructions, demonstrations, and options for the same task.  
Given context $X$, we first acquire the skill neurons $\mathcal{N}_s^X$ and the accuracy $\text{ACC}_{\mathcal{N}_s^X}^X$. 
Then, we use the classifier built with $\mathcal{N}_s^X$ to evaluate the task by context $Y$  as $\text{ACC}_{\mathcal{N}_s^X}^Y$. 
We denote the robustness of $\mathcal{N}_s^X$ on context $Y$ as $\frac{\max(\text{ACC}_{\mathcal{N}_s^X}^Y-\alpha, 0)}{\max(\text{ACC}_{\mathcal{N}_s^Y}^Y-\alpha, 0)}$, where $\alpha$ is the accuracy by Rand.}

\cmr{Using PAWS as an example, we create 12 distinct contexts by varying the instructions, the selection of demonstrations, and the output token styles. 
By measuring robustness across combinations, we find that it remains very high (near 1) for varying instructions and demonstrations but drops significantly when target tokens change.
The results show that skill neurons remain highly robust to changes in instructions and demonstrations but lose robustness when output tokens change.
See \S~\ref{sec:appendix-generality-instructions} for details of experimental settings and results.}

% \subsection{Inclusivity}
\subsection{Are Neurons Substitutable in Representing Language Skills?}
\label{sec:neuron-inclusivity}
\cmr{We investigate whether skill neurons uniquely represent skills or can be substituted by different neurons. 
For investigation, we build Magn-Probes using different neuron sets.
Specifically, we select 64 consecutive neurons from the ranked list, ordered by their importance as skill indicators (\S~\ref{para:magnitude-probe}).\footnote{We use 64-neuron units that maintain high accuracy across tasks (\S~\ref{sec:appendix-neurongen}). 
With 352,256 neurons in Llama2-7B’s FF layers, this produces 5,504 unique neuron sets.}}

\begin{figure}[t]
    \centering
    \includegraphics[width=0.97\linewidth]{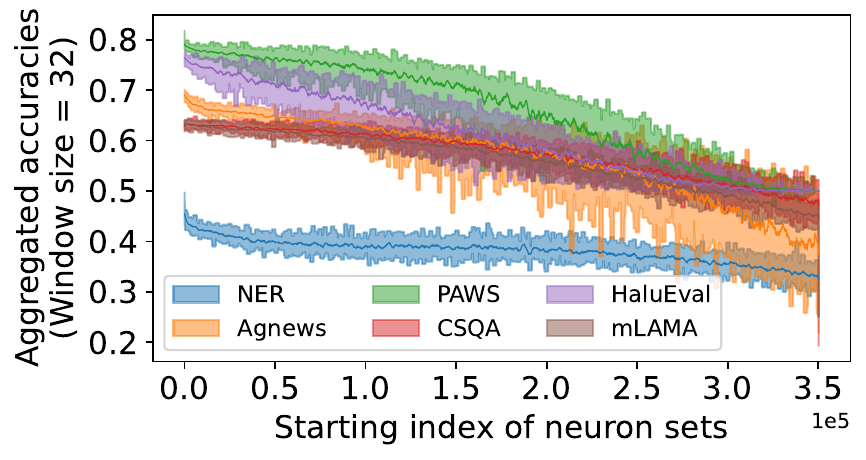}
    \caption{Accuracies of Magn-Probe with different neuron sets, plotting the mean accuracy within each window, along with the accuracy ranges, as the envelope. 
    Neuron sets are selected from all neurons in Llama2-7B in groups of 32, ranked by importance as skill indicators.}
    \label{fig:inclusivity}
\end{figure}

\cmr{Figure~\ref{fig:inclusivity} depicts the accuracies across six tasks with different neuron sets, showing that representational accuracy declines only gradually with less important neurons.
Even using the least important neurons still outperforms Rand.
This suggests that numerous neurons can act as skill indicators and skill neurons are broadly distributed and substitutable (see \S~\ref{sec:appendix-neuron-set} for full results).}

\begin{figure}[t]
    \centering
    \includegraphics[width=0.97\linewidth]{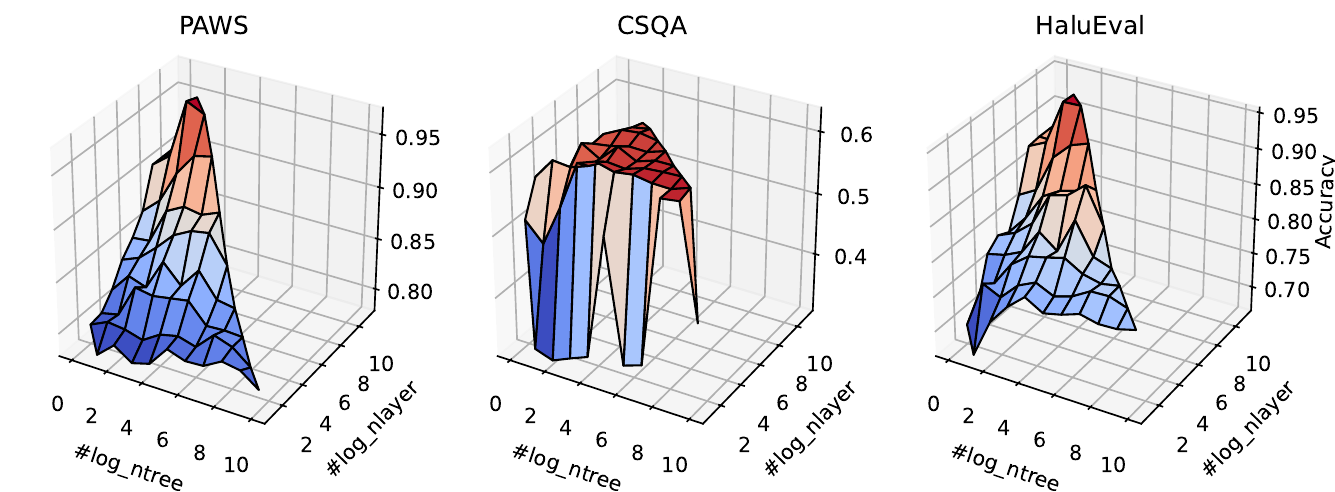}
    \caption{Accuracies of Tree-Probes with varying depths and trees. \textit{X-axis}: logarithm of trees' number; \textit{Y-axis}: logarithm of tree depths; \textit{Z-axis}: Accuracy.}
    \label{fig:tree_specs_few-shot}
\end{figure}

\subsection{Do Skill Neurons Depend on Each Other?}
\label{sec:neuron-hierarchy}
\cmr{The majority-vote probes assume neuron independence, while the Tree-Probe model their interdependencies via hierarchical classifiers. 
Figure~\ref{fig:full_acc_rader} shows that Tree-Probles better represents language skills by accounting for neuron interdependency. 
To see how important interdependency is in encoding language skills, we train Tree-Probes with varying \texttt{\#n\_trees} and \texttt{\#n\_layers}.\footnote{\texttt{\#n\_trees} and \texttt{\#n\_layers} are set to $2^N$ and $2^M$, where  $0 \leq N \leq 10$, $1 \leq M \leq 11$, and $N + M < 12$.}}

\cmr{Figure~\ref{fig:tree_specs_few-shot} reports the resulting accuracies of Tree-Probe on PAWS, CSQA, and HaluEval. 
Their different shapes indicate the interdependency style between neurons for diverse skills differs. 
Some tasks (PAWS) prefer deep layers, while some (CSQA) prefer more trees, and some (HaluEval) require a balance.
See \S~\ref{sec:appendix-neuron-hierarchy} for details.}

\section{Related Work}

\smallskip\noindent\textbf{Neuron-level knowledge attribution methods.}
Previous work links neurons to knowledge~\cite{oba-etal-2021-exploratory} by measuring their impact on model predictions.
Some use causal interventions~\cite{geva-etal-2021-transformer,meng2022locating,chen2023journeycenterknowledgeneurons,wang-etal-2024-unveiling}, while others rely on heavy tensor calculations~\cite{geva-etal-2022-transformer, lee2024mechanisticunderstandingalignmentalgorithms, yu-ananiadou-2024-neuron}.
These methods are computationally costly, limiting their use for large-scale probing across diverse LLM prompts. 
They also focus on relative neuron rankings rather than the precise, quantitative neuron-output relationships, reducing their applicability in use cases like knowledge editing~\cite{zhang-etal-2024-knowledge-editing} and bias mitigation~\cite{gallegos-etal-2024-bias}.
Gradient-based approaches~\citep{lundstrom2022rigorousstudyintegratedgradients, dai-etal-2022-knowledge} similarly face high computational costs.

% They drive the development of knowledge attribution methods that assign importance scores to groups of features, indicating their relevance to the model output for a given input, including gradient-based method~\citep{dai-etal-2022-knowledge, sundararajan2017axiomaticattributiondeepnetworks}, casual intervention methods that modify the internal status of models and observe the causal effect~\citep{meng2022locating, goldowskydill2023localizingmodelbehaviorpath} and automatic-tool-based methods relying on self-explanation with LLMs~\citep{conmy2023automatedcircuitdiscoverymechanistic,singh2023explainingblackboxtext}.
% While these methods offer valuable insights into the interpretability of LLMs, such as neuron-ngram~\citep{voita-etal-2024-neurons} or neuron-fact connections~\citep{dai-etal-2022-knowledge}, they provide only qualitative measures of neuron importance, leaving the quantitative relationship between neurons and model output unexplored. 
% Our research discovers the linear relationship between neuron and output, providing a numerical estimation of their relation. 

\smallskip\noindent\textbf{Skill neuron probing.}
Neurons in feed-forward layers can encode specific language skills, enabling the tasks to be solved using only their activations; such neurons are called skill neurons~\cite{wang-etal-2022-finding-skill, song-etal-2024-large}. 
Prior work showed that neurons represent semantic skills like sentiment classification~\cite{wang-etal-2022-finding-skill, song-etal-2024-large} and complex skills such as style transfer~\cite{lai-etal-2024-style} and translation~\cite{tan-etal-2024-neuron}. 
However, these studies focus on activations as knowledge indicators, overlooking neuron gradients' potential to represent language skills, which limits neuron-level model adjustments.
% highlighting their representational ability but ignoring their limited influence on model output. In contrast, our findings about NEG provide a stronger basis for knowledge control.

\section{Conclusions}

% Through extensive neuron intervention experiments, 
This is the first study to provide a global quantitative link between feed-forward layer neurons and model output, and discuss the potential of precise LLM control via neuron modifications.
Through neuron intervention experiments, we reveal the linear relationship between neurons and model outputs and quantify such linearity by ``neuron empirical gradients'' (NEG) and propose NeurGrad, an efficient NEG estimation method. 
Finally, we verify NEG's ability to represent various language skills associated with diverse prompts through skill neuron probing. 

As future work, we plan first to explore the trade-off between precision and intensity in neuron modification based on NeurGrad, and then leverage these insights to develop a broader neuron-level modification methodology for applications such as knowledge editing and bias mitigation.

\section{Limitations}
Our research establishes a framework to quantitatively measure neuron influence on model output and shows that empirical gradients effectively represent language skills, linking language skill representation to outputs via neuron empirical gradients.
However, tuning neuron values for skill-level output adjustment remains unexplored and could offer a more efficient alternative to weight-level tuning. 
We also plan to assess NEG's role in language generation skills, aiming for dynamic behavior changes without altering LLM parameters, reducing costs and improving flexibility in model adaptation.

\shin{Currently, our neuron linearity and gradient analysis focus on single-token factual prompts. Future work will expand to diverse domains and 
% investigate neuron attribution methods for 
multi-token contexts to support wider use cases with language generation.}

\section*{Acknowledgments}
This paper is based on results obtained from a project, JPNP22007, commissioned by the New Energy and Industrial Technology Development Organization (NEDO).

\bibliography{acl_latex}

\appendix

\section{Generality of Neuron Linearity }
% \label{sec:appendix-neuron-linearity-analysis}
\label{sec:appendix-neurongen}

In this section, we provide additional evidence to verify that linearity is a general property for neurons in LLMs. 
Specifically, we want to verify whether the linear neurons exist widely across different Transformer feed-forward layers and within different prompts. We use the metrics of layer generality (\textbf{LG}) and 
prompt generality (\textbf{PG}) to measure the prevalence of their existence. Intuitively, we can consider a simplified problem as follows: suppose we have many colored balls (green, blue, ...) and 10 bins, and if we want to verify whether the blue ball has ``generality,'' it means (1) \textbf{high coverage}: the blue ball exists in most of the bins; (2) \textbf{even distribution}: the number of blue balls in each bin hardly differs from others. 
For our neuron generality, the ``balls'' are the ``linear neurons,'' and the ``bins'' refer to either ``feed-forward layers'' (for \textbf{LG}) or ``different prompt'' (for \textbf{PG}). To address these two aspects simultaneously, we define \textbf{LG} and \textbf{PG} as follows:

\begin{equation}
    \textbf{LG} \triangleq \text{coverage}_\mathrm{layer} \times \text{distribution}_\mathrm{layer},
\end{equation}

\begin{equation}
    \textbf{PG} \triangleq \text{coverage}_\mathrm{prompt} \times \text{distribution}_\mathrm{prompt},
\end{equation}

where $\text{coverage}$ and $\text{distribution}$ are defined as:
\begin{equation}
    \text{coverage}_{x} = \frac{\Sigma_{i} \mathds{1}(\mathrm{linear\ neuron\ exists\ in\ } x_{i})}{\mathrm{\#\ of\ }x},
\end{equation}

\begin{equation}
    \text{distribution}_{x} = 1 - \frac{\mathrm{Var}(\mathrm{\#neurons\ in\ }x)}{\mathrm{maxVar(\#neurons\ in\ }x)},
\end{equation}
where $x$ refers to either layer or prompt, $\mathrm{maxVar}(\cdot)$ denotes the max possible variance. High coverage and distribution are desirable; a perfect generality then achieves coverage of one and distribution of one. 
The statistics in Table~\ref{tab:general_neuron_linearity} suggest that linearity is a general property of neurons, largely independent of specific prompts or Transformer layers.

\begin{table}[t]
\centering
\small
\tabcolsep 5.5pt
\begin{tabular}{lrrr}
\toprule
& \textbf{\makecell{Linear\\neuron\\ratio}} & \textbf{\makecell{Prompt-\\wise\\gen.}} & \textbf{\makecell{Layer-\\wise\\gen.}} \\\midrule
BERT$_\mathrm{base}$   & .9565 & .9999 & .9982 \\
BERT$_\mathrm{large}$  & .8756 & .9999 & .9989 \\
% BERT$_\mathrm{wwm}$    & .9564 & .9999 & .9990 \\
Llama2-7B              & .9387 & .9999 & .9986 \\
% Llama2-13B             & .9677 & .9999 & .8618 \\
Llama2-70B             & .9208 & .9999 & .6294 \\
\bottomrule
\end{tabular}
\caption{Neuron linearity statistics. We choose 1000 prompts and their corresponding 100 neurons with top gradient magnitudes. 
For Llama2-70B, since the model is giant, we only chose 200 prompts and 100 neurons due to the high computational cost. The shift range is set to $\pm$ 2.}
\label{tab:general_neuron_linearity}
\end{table}

\section{Neurons' Statistics}

\subsection{Distribution of Neuron Activations}
\label{sec:appendix-activation-distribution}

\begin{figure*}[t]
    \centering
    \includegraphics[width=1\linewidth]{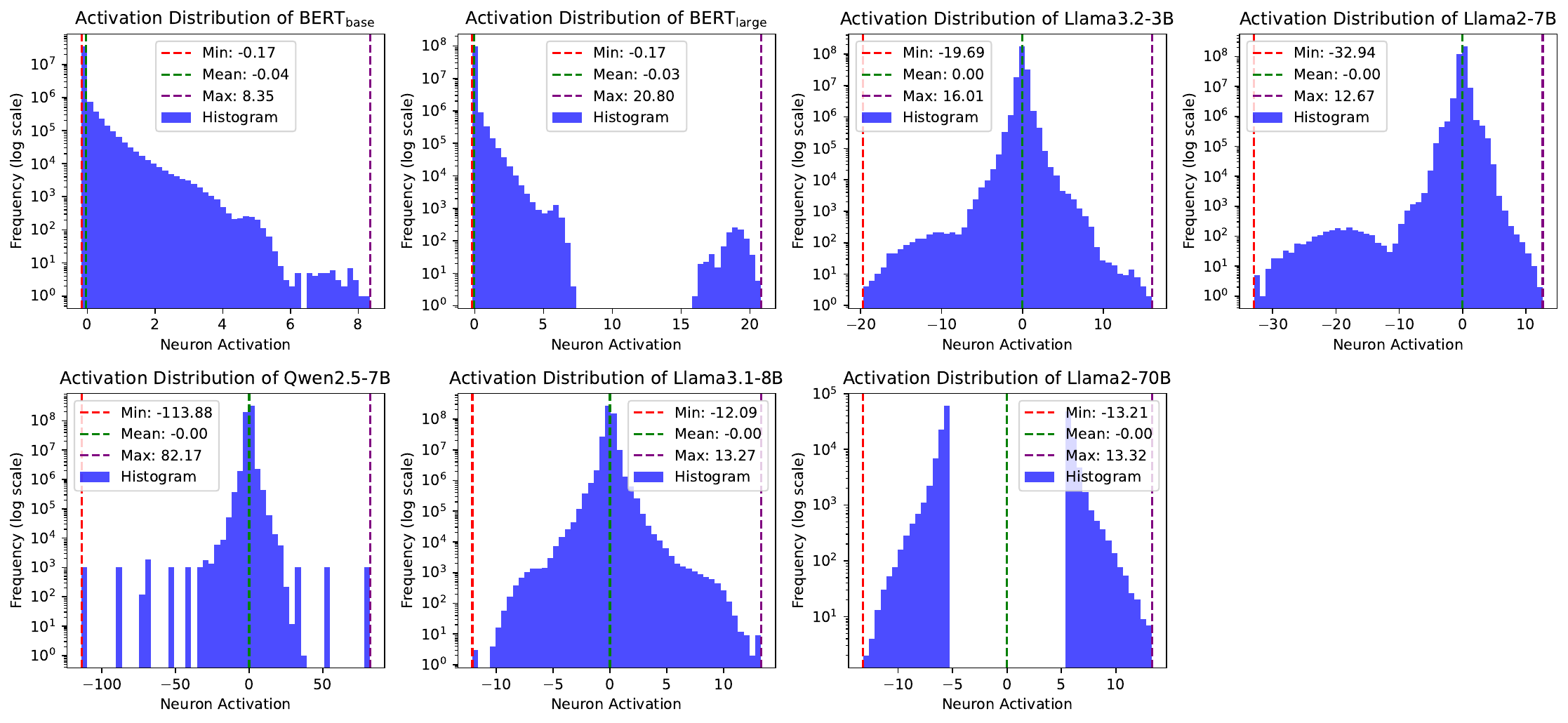}
    \caption{Histograms of neuron activations for seven models across 1,000 prompts, displayed on a logarithmic y-axis. The figure includes three BERT models and three LLaMA-2 models, with each subplot showing the distribution of activations for one model.}
    \label{fig:activation-distributions}
\end{figure*}

In this section, we analyze the distribution of neuron activations across seven PLMs, illustrated in Figure~\ref{fig:activation-distributions}. The models include three BERT-based PLMs and three instruction-tuned LLaMA-2 LLMs. Figure~\ref{fig:activation-distributions} reveals that most neuron activations fall within the range of $\pm$10. 
While there are still some neurons that have a value out of the range of $\pm$10, the number of such neurons is comparably fewer, and increasing the range linearly increases the computational cost. 
Considering the balance between coverage and computational cost, we finally set the intervention range as $\pm$10 as shown in Section \S~\ref{sec:neuron_analysis}.

\subsection{Distribution of Neuron NEGs}
\label{sec:appendix-gradient-distribution}

\begin{figure*}[t]
    \centering
    \includegraphics[width=1\linewidth]{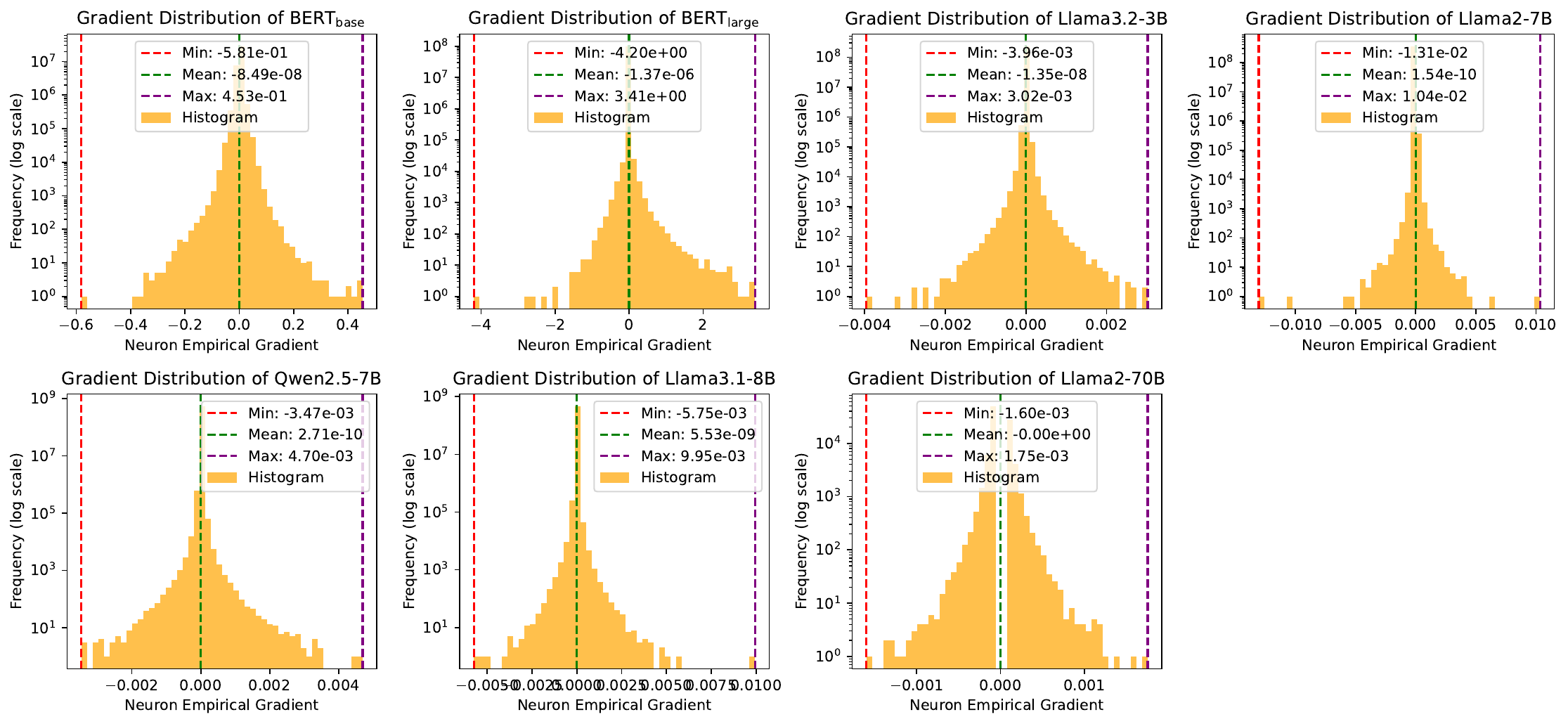}
    \caption{Histograms of neuron NEGs for seven models across 1,000 prompts, displayed on a logarithmic y-axis.}
    \label{fig:gradient-distributions}
\end{figure*}

In this section, we report the distribution of neurons' NEG across the seven PLMs, as illustrated in Figure~\ref{fig:gradient-distributions}. Similar to our discussion in Section \S~\ref{sec:neg-distributions}, neurons capable of altering model output are not rare. 
For instance, in BERT$_\mathrm{base}$, over 1,000 neurons exhibit NEG magnitudes larger than 0.1. 
Additionally, the NEG magnitudes of neurons in LLaMA-2 LLMs are significantly smaller than those in BERT PLMs, likely due to the smaller parameter size of BERT models, which grants individual neurons greater influence over model output. 
Notably, all models exhibit zero gradients when averaging the NEGs across all neurons.

\subsection{How are neurons distributed across layers?}
\label{sec:appendix-neuron-layer-distribution}

\begin{figure*}[t]
    \centering
    \includegraphics[width=1\linewidth]{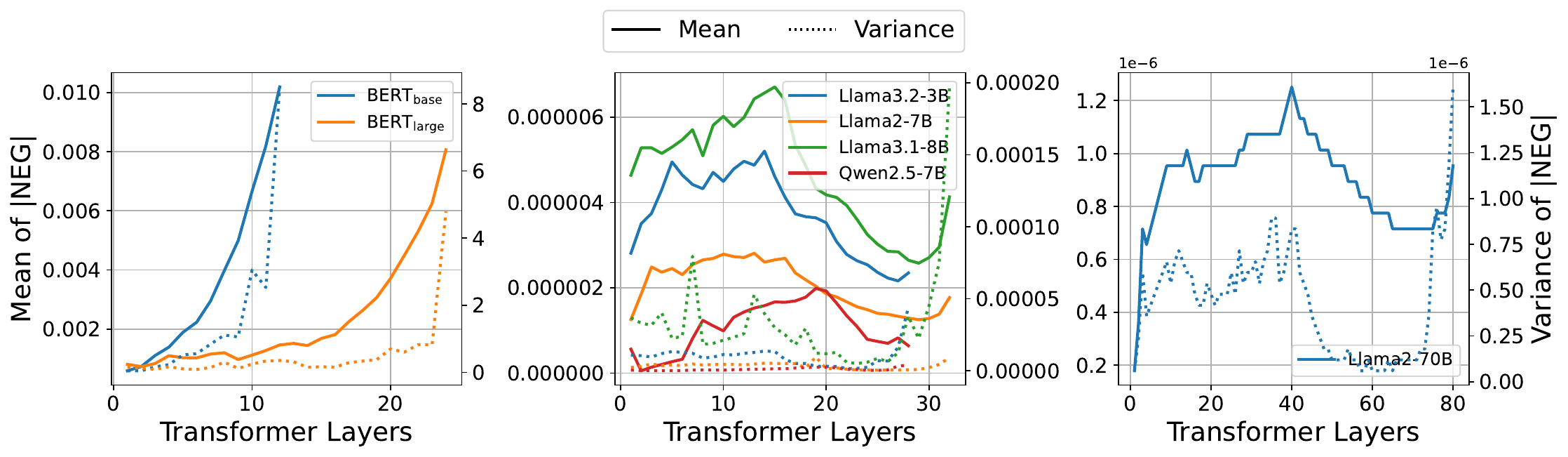}
    \caption{Means and variances of NEG magnitudes across Transformer layers on seven models.
    The data is calculated from the average of 1000 prompts.}
    \label{fig:grad_layer_distri}
\end{figure*}

We examine the variation in the absolute value of NEGs across Transformer layers to understand the distribution of neuron controllability.
The Figure~\ref{fig:grad_layer_distri} illustrates the means and confidence interval of magnitudes of NEG across layers. 
The mean NEG magnitude reflects the intensity with which PLMs adjust output probabilities through neurons in a given layer, while the variance indicates how concentrated the effective neurons are within that layer.
A positive Corr is observed between variances and means,\footnote{The Corr between means and variances of neuron magnitudes across different layers are 0.88, 0.79, 0.87 for \textbf{BERT$_\mathrm{base/large/wwm}$}, and 0.57, 0.51, 0.42 for \textbf{Llama2-7B/13B/70B}.}
suggesting that as PLMs increase the intensity of gradient activity in specific layers, they also focus more on a limited subset of neurons. 
% In the shallow layers, gradients are generally low and dispersed across neurons, indicating that these layers perform broad linguistic processing, with many neurons contributing modestly to the final predictions. 
% In contrast, deeper layers show increased gradient magnitudes and higher concentrations, reflecting more targeted and confident adjustments by the model.
Specifically, in BERT models, a strong Corr is evident between layer depth and neuron controllability intensity, with deeper layers exhibiting larger gradient magnitudes, whereas Llama2 displays a distinct pattern: gradient magnitudes peak in the middle layers, decrease towards the deeper layers and then increase at the final layers.
This divergence underscores the differences between the BERT and Llama2 families, emphasizing the need for case-by-case analysis in LLM mechanism investigation.

\begin{figure}[t]
    \centering
    \begin{subfigure}[b]{0.98\linewidth}
        \includegraphics[width=\linewidth]{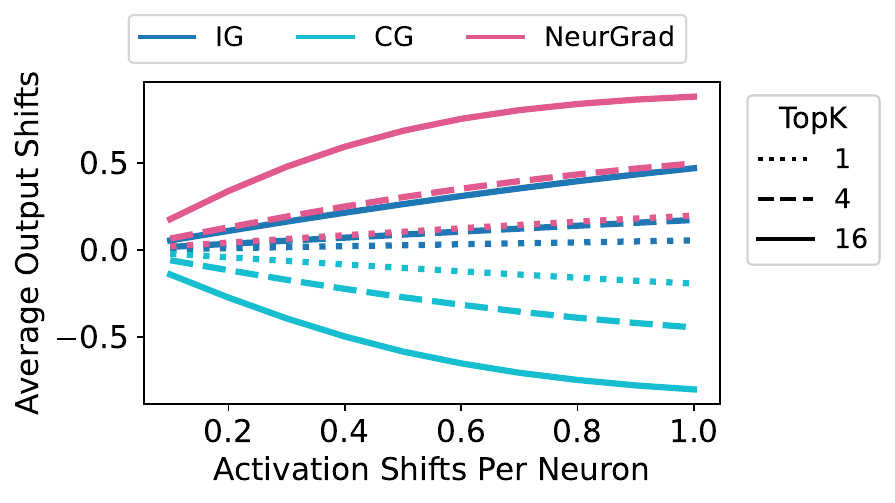}
        \caption{Evaluation results on BERT$_\mathrm{base}$.}
        \label{fig:kae-bert-base}
    \end{subfigure}
    \hfill
    \begin{subfigure}[b]{0.98\linewidth}
        \includegraphics[width=\linewidth]{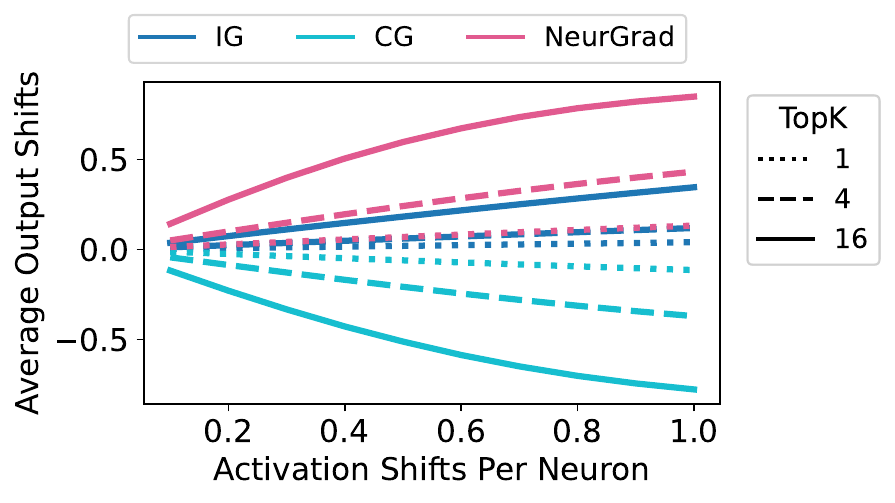}
        \caption{Evaluation results on BERT$_\mathrm{large}$.}
        \label{fig:kae-bert-large}
    \end{subfigure}
    % \hfill
    % \begin{subfigure}[b]{0.98\linewidth}
    %     \includegraphics[width=\linewidth]{figures/attribution_method_comparision_bert-wwm.pdf}
    %     \caption{Evaluation results on BERT$_\mathrm{wwm}$.}
    %     \label{fig:kae-bert-wwm}
    % \end{subfigure}
    \caption{Knowledge attribution evaluation by comparing CG, IG, and NeurGrad on two BERT.}
    \label{fig:knowledge-attribution-evaluation-combined}
\end{figure}

\section{Knowledge Attribution Evaluation: Supplementary Experiments}
\label{sec:appendix-supplementary-knowledge-attribution}
In this section, we report the knowledge evaluation experiments on other PLMs, including two BERT PLMs. 
We exclude LPI from the following experiments as LPI cannot be applied to BERT models. 
We follow a similar experiment setup to Section \S~\ref{sec:knowledge-attribution-evaluation}. 

The evaluation results are illustrated in Figure~\ref{fig:knowledge-attribution-evaluation-combined}. 
NeurGrad consistently outperforms other gradient-based methods in finding the top-$K$ important neurons.
Furthermore, we can observe that output shifts made on BERT PLMs are much larger than shifts on Llama2-7B (Figure~\ref{fig:attribution_method_comparision}). 
This is due to the small NEG magnitudes in Llama2-7B as introduced in \S~\ref{sec:appendix-gradient-distribution}.

\begin{figure*}[t]
    \centering
    \begin{subfigure}[b]{0.48\linewidth}
        \includegraphics[width=\linewidth]{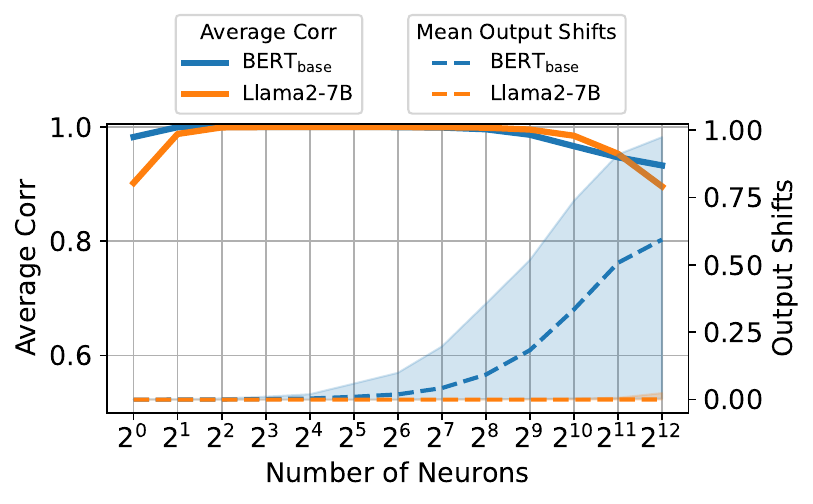}
        \caption{Enhancement range of [0, 0.1].}
        \label{fig:multi-neuron-0.1}
    \end{subfigure}
    \hfill
    \begin{subfigure}[b]{0.48\linewidth}
        \includegraphics[width=\linewidth]{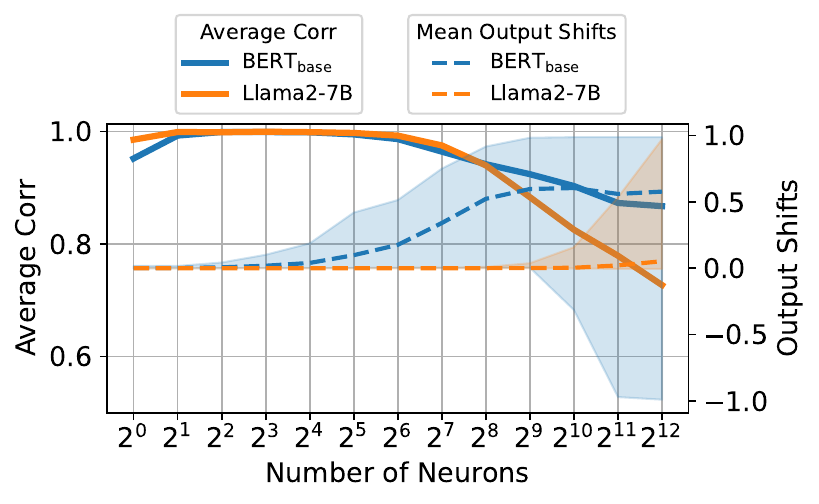}
        \caption{Enhancement range of [0, 1].}
        \label{fig:multi-neuron-1}
    \end{subfigure}
    
    \vspace{1cm} % Adjust vertical spacing between rows
    
    \begin{subfigure}[b]{0.48\linewidth}
        \includegraphics[width=\linewidth]{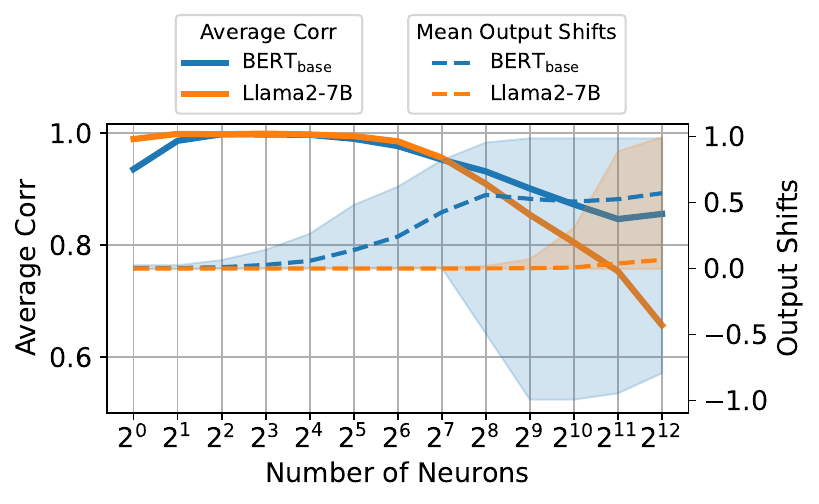}
        \caption{Enhancement range of [0, 1.5].}
        \label{fig:multi-neuron-1.5}
    \end{subfigure}
    \hfill
    \begin{subfigure}[b]{0.48\linewidth}
        \includegraphics[width=\linewidth]{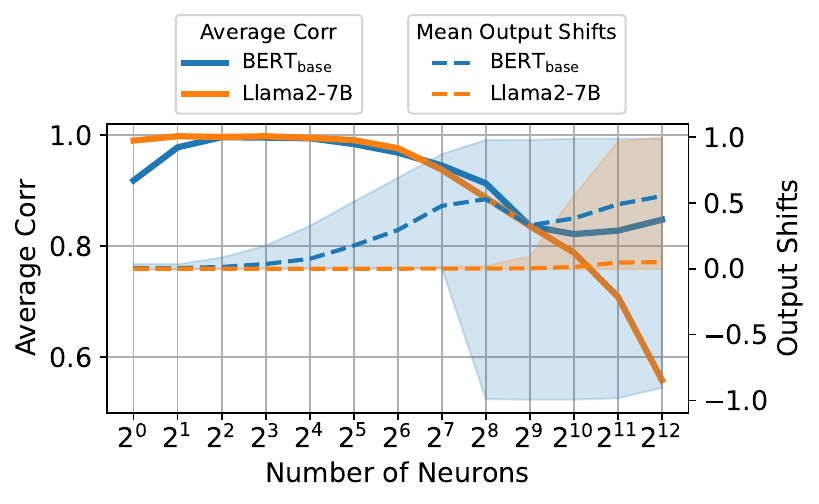}
        \caption{Enhancement range of [0, 2].}
        \label{fig:multi-neuron-2}
    \end{subfigure}
    
    \caption{Multi-neuron intervention experiment results with different enhancement ranges.}
    \label{fig:multi-neuron-combined}
\end{figure*}

\section{Multi-neuron Intervention: Supplementary Experiments}
\label{sec:appendix-supplementary-multineuron-intervention}

In this section, we report the multi-neuron intervention experiments conducted with different enhancement ranges on BERT$_\mathrm{base}$ and Llama2-7B, following the similar experiment setup to Section \S~\ref{sec:multineuron-intervention}. 
Specifically, we report the correlation between output shift and the accumulated NEGs estimated by NeurGrad with the enhancement ranges of [0, 0.1], [0, 1], [0, 1.5], and [0, 2], illustrated in Figure~\ref{fig:multi-neuron-combined}.
Figure~\ref{fig:multi-neuron-combined} demonstrates that a larger enhancement range, involving more neurons, can vastly reduce the Corr, suggesting the output shift is less predictable. 
However, we can observe that BERT$_\mathrm{base}$ consistently achieves strong Corr. (>0.8) for any scenario. 
While Llama2-7B is less stable than BERT$_\mathrm{base}$, it can still maintain moderate positive Corr. (>0.5) for 4096 neurons with an enhancement range of [0,2].

% , 1 (Figure~\ref{fig:multi-neuron-1}), 1.5 (Figure~\ref{fig:multi-neuron-1.5}) and 2 (Figure~\ref{fig:multi-neuron-2}).

\section{Model cards}
\label{sec:appendix-mode_cards}
Here are the links from Hugging Face to load each model:
\begin{description}
    \item[\small BERT$_\mathrm{base}$:] 
    \small \url{https://huggingface.co/bert-base-uncased}
    \item[\small BERT$_\mathrm{large}$:] 
    \small \url{https://huggingface.co/bert-large-uncased}
    % \item[\small BERT$_\mathrm{wwm}$:] 
    % \small \url{https://huggingface.co/bert-large-uncased-whole-word-masking}
    \item[\small Llama3.2-3B:] 
    \small \url{https://huggingface.co/meta-llama/Llama-3.2-3B-Instruct}
    \item[\small Llama2-7B:] 
    \small \url{https://huggingface.co/meta-llama/Llama-2-7B-hf} 
    \item[\small Qwen2.5-7B:] 
    \small \url{https://huggingface.co/Qwen/Qwen2.5-7B-Instruct} 
    \item[\small Llama3.1-8B:] 
    \small \url{https://huggingface.co/meta-llama/Llama-3.1-8B-Instruct}
    \item[\small Llama2-70B:] 
    \small \url{https://huggingface.co/meta-llama/Llama-2-70B-hf} 
    
\end{description}

\begin{table}[t]
\centering
\small
\begin{tabular}{lcc}
\toprule
\textbf{Model}        & \textbf{\#n\_layers} & \textbf{\#neurons\_per\_layer} \\
\midrule
BERT$_\mathrm{base}$  & 12 & 3,072 \\
% BERT$_\mathrm{large}$ & 24 & 4,096 \\
BERT$_\mathrm{wwm}$   & 24 & 4,096 \\
Llama3.2-3B           & 28 & 11,008 \\
Llama2-7B             & 32 & 11,008 \\
Qwen2.5-7B            & 28 & 18,944 \\
Llama3.1-8B           & 32 & 14,336 \\
Llama2-70B            & 80 & 28,672\\
\bottomrule
\end{tabular}
\caption{Number of Layers and Intermediate Neurons per Layer for BERT and Llama2 Models}
\label{tab:model_neurons}
\end{table}
The statistics of these seven PLMs, including the number of layers (\#n\_layers) and neurons per layer (\#neurons\_per\_layer) are listed in Table~\ref{tab:model_neurons}. 

\section{Construction of MCEval8K}
\label{sec:appendix-mceval8k}

The motivation behind creating MCEval8K is to establish a comprehensive benchmark that spans diverse knowledge genres and language skills. 
Since we aim to facilitate skill neuron probing experiments where a single token must represent answers, we adopt a multi-choice task format. 
Additionally, we aim for the benchmark to be adaptable while avoiding redundancy for effective evaluation. In summary, we adhere to several guiding principles to design MCEval8K.

\begin{enumerate}
\item All datasets must be in multi-choice format. 
\item Avoid including datasets that convey similar language skills.
\item To eliminate potential bias from imbalanced classifications, we ensure that the number of correct options is evenly distributed across all answer choices. This balance helps maintain fairness and accuracy in the analysis results.
\item We use a unified number (8000) of data to avoid high computational costs. 
\end{enumerate}

\smallskip\noindent\textbf{Multi-choice format:} We created MCEval8K to include six different genres with 22 tasks, which are linguistic, content classification, natural language inference (NLI), factuality, self-reflection, and multilingualism. 
All the genres and tasks are listed in Table~\ref{tab:dataset_details}. 
For datasets that are not multi-choice tasks, we create options for each inquiry following rules. 
These datasets include POS, CHUNK, NER, MyriadLAMA, TempLAMA, Stereoset, M-POS, and mLAMA. 
The rules we adhere to create options are listed below:

\begin{description}
    \item[POS] We use weighted sampling across all POS tags to select three additional tags alongside the ground-truth tag.
    \item[CHUNK] The process is analogous to POS.
    \item[NER] The process is analogous to POS.
    \item[MyriadLAMA] For factual inquiries formed from $<\text{sub}_i, \text{rel}_j>$, we collect all objects that appear as the target of the $\text{rel}_j$ within the dataset and perform sampling to select three additional objects alongside the ground-truth tag.
    \item[TempLAMA] We randomly sample three additional candidate years from the range 2009 to 2020, alongside the ground-truth tag.
    \item[M-POS] The process is similar to POS, applied separately for each language.
    \item[mLAMA] The process is similar to MyriadLAMA, applied separately for each language.
\end{description}

\smallskip\noindent\textbf{Balanced Options:} Most datasets, except for Stereoset, contain over 8000 data points. To ensure balance across all options, we perform balanced sampling so that each option has an equal number of examples. 
From these datasets, we split 8000 examples into training, validation, and test sets, allocating 6,000, 1000, and 1000 examples, respectively. For instance, in the case of mLAMA, where each inquiry has four options, we ensure that the correct answer is represented equally across all four positions. This results in 1,500 occurrences (6,000/4) per position in the training set and 250 occurrences per position in both the validation and test sets.

\smallskip\noindent\textbf{Creation of multilingual tasks:} For multilingual datasets, we focus on five languages: English (en), German (de), Spanish (es), French (fr), and Chinese (zh). These languages vary significantly in linguistic distance, with English being closer to German, French closer to Spanish, and Chinese being distant from all of them. 
This selection allows for a deeper analysis considering linguistic distances between languages.
We ensure that 5 languages have the same number of data examples in each dataset (1,600 per language). 
Furthermore, for datasets like mLAMA, XNLI, and M-AMAZON, we ensure that each piece of knowledge is expressed in all five languages. This consistency enables direct comparisons of language understanding abilities across different languages.

\begin{table*}[t]
\centering
\small
\setlength{\tabcolsep}{6pt} % Increases column spacing
\begin{tabular}{p{2cm}lp{3.4cm}p{3.9cm}cc}
\toprule
\textbf{Genres} & \textbf{Task} & \textbf{Language skills} & \textbf{Dataset} & \textbf{\#n\_choices} & \textbf{\#n\_examples} \\
\midrule

\multirow{8}{*}[-7pt]{Linguistics}  
& POS   & Part-of-speech tagging &\href{https://github.com/UniversalDependencies/UD_English-GUM}{Universal Dependencies}~\citep{ud}       & 4 & 8000 \\\cmidrule{2-6}
& CHUNK & Phrase chunking &\href{https://huggingface.co/datasets/eriktks/conll2000}{CoNLL-2000}~\citep{conll2000}           & 4 & 8000 \\\cmidrule{2-6}
& NER   & Named entity recognition &\href{https://huggingface.co/datasets/eriktks/conll2003}{CoNLL-2003}~\citep{conll2003}           & 4 & 8000 \\\cmidrule{2-6}
& GED   & Grammatic error detection  &\href{https://github.com/google-research-datasets/clang8}{cLang-8}~\citep{clang8-1,clang8-2}  & 2 & 8000 \\
\midrule

\multirow{4}{*}[-4pt]{\makecell[l]{Content\\classification}} 
& IMDB    & Sentiment classification &\href{https://huggingface.co/datasets/stanfordnlp/imdb}{IMDB}~\citep{imdb}               & 2 & 8000 \\\cmidrule{2-6}
& Agnews  & Topic classification &\href{https://huggingface.co/datasets/fancyzhx/ag_news}{Agnews}~\citep{agnews}          & 4 & 8000 \\\cmidrule{2-6}
& Amazon  & Numerical sentiment classification &\href{https://huggingface.co/datasets/McAuley-Lab/Amazon-Reviews-2023}{Amazon Reviews}~\citep{amazon-review} & 5 & 8000 \\
\midrule

\multirow{4}{*}[-4pt]{\parbox[t]{2cm}{\makecell[l]{Natural\\language\\inference (NLI)}}} 
& MNLI    & Entailment inference &\href{https://huggingface.co/datasets/nyu-mll/glue/viewer/mnli}{MNLI} ~\citep{mnli} & 3 & 8000 \\\cmidrule{2-6}
& PAWS    & Paraphrase identification &\href{https://huggingface.co/datasets/google-research-datasets/paws}{PAWS} ~\citep{paws}          & 2 & 8000 \\\cmidrule{2-6}
& SWAG  & Grounded commonsense inference &\href{https://huggingface.co/datasets/allenai/swag}{SWAG} ~\citep{swag} & 4 & 8000 \\
\midrule

\multirow{7}{*}[-8pt]{\parbox[t]{2cm}{Factuality}} 
& FEVER    & Fact checking &\href{https://huggingface.co/datasets/fever/fever}{FEVER} ~\citep{Fever} & 2 & 8000 \\\cmidrule{2-6}
& MyriadLAMA & Factual knowledge question-answering &\href{https://huggingface.co/datasets/iszhaoxin/MyriadLAMA}{MyriadLAMA} ~\citep{zhao-etal-2024-matters} & 4 & 8000 \\\cmidrule{2-6}
& CSQA  & Commonsense knowledge question-answering &\href{https://huggingface.co/datasets/tau/commonsense_qa}{CommonsenseQA} ~\citep{commonsenseqa} & 4 & 8000 \\\cmidrule{2-6}
& TempLAMA  & \parbox[t]{3.5cm}{Temporary facts question-answering} &\href{https://huggingface.co/datasets/Yova/templama}{TempLAMA} ~\citep{templama} & 4 & 8000 \\
\midrule

\multirow{4}{*}[-4pt]{\makecell[l]{Self-reflection}} 
& HaluEval    & Hallucination detection &\href{https://huggingface.co/datasets/pminervini/HaluEval/viewer/dialogue_samples}{HaluEval-diag} ~\citep{halueval} & 2 & 8000 \\\cmidrule{2-6}
& Toxic    & Toxicity post identification &\href{https://huggingface.co/datasets/google/jigsaw_toxicity_pred}{Toxicity prediction} ~\citep{toxic}          & 2 & 8000 \\\cmidrule{2-6}
& Stereoset  & Social stereotype detection &\href{https://huggingface.co/datasets/McGill-NLP/stereoset}{Stereoset} ~\citep{stereoset} & 3 & 4230 \\
\midrule

\multirow{10}{*}[-10pt]{\makecell[l]{Multilinguality}} 
& LTI    & Language identification &\href{https://huggingface.co/datasets/SEACrowd/lti_langid_corpus}{LTI LangID corpus} ~\citep{lti-1,lti-2} & 5 & 8000 \\\cmidrule{2-6}
& M-POS    & Multilingual POS-tagging &\href{https://github.com/universaldependencies}{Universal Dependencies} ~\citep{ud}          & 4 & 8000 \\\cmidrule{2-6}
& M-Amazon  & Multilingual Amazon review classification &\href{https://www.kaggle.com/datasets/mexwell/amazon-reviews-multi}{Amazon Reviews Multi} ~\citep{marc_reviews} & 5 & 8000 \\\cmidrule{2-6}
& mLAMA  & Multilingual factual knowledge question-answering &\href{https://huggingface.co/datasets/cis-lmu/m_lama}{mLAMA} ~\citep{mlama} & 4 & 8000 \\\cmidrule{2-6}
& XNLI  & Multilingual entailment inference &\href{https://huggingface.co/datasets/facebook/xnli}{XNLI} ~\citep{xnli} & 3 & 8000 \\
\bottomrule
\end{tabular}
\caption{Details of datasets in MCEval8K.}
\label{tab:dataset_details}
\end{table*}

\section{Details of Skill Neuron Probing}
\subsection{Per-task Probing Result}
\label{sec:appendix-per-task-result}

In this section, we report the details of our skill neuron probing evaluation, including the full optimal accuracies on all tasks with zero-shot prompt setting (Table~\ref{tab:per-task-acc-zeroshot}) and few-shot prompt setting (Table~\ref{tab:per-task-acc-fewshot}). 
For two majority vote probes, optimal accuracies are acquired by performing a hyperparameter (optimal neuron size) search on the validation set and evaluating the test set. 
We report the optimal neuron sizes for all tasks and the table's accuracies.
For the random-forest probing (Tree-Probe), we directly use the gradients of all neurons to train the random forest tree. 
As the random forest training algorithm only takes important features to construct the decision trees, we also report the number of neurons used to build the random forests. 
The details of the random-forest-based probe are introduced in \S~\ref{sec:appendix-random-forest-probe}.

\begin{table*}[t]
    \centering
    \normalsize
    \setlength{\tabcolsep}{5.5pt}
    \resizebox{0.8\textwidth}{!}{
    \begin{tabular}{llllll}
    \toprule
    Tasks & \textbf{\makecell[c]{Rand}} & \textbf{\makecell[c]{LM-Prob}} & \textbf{
\makecell[c]{Polar-Probe\\(\#n\_neurons)}} & \textbf{\makecell[c]{Magn-Probe\\(\#n\_neurons)}} & \textbf{\makecell[c]{Tree-Probe\\(\#n\_neurons)}} \\ \midrule
    GED          & .5000 & .5000 & .7580 (16) & .8050 (1024) & 1.000 (54644) \\
    POS          & .2500 & .5050 & .5190 (16) & .5470 (4) & .5850 (91290) \\
    CHUNK        & .2500 & .3510 & .4660 (8) & .4490 (16) & 1.000 (93282) \\
    NER          & .2500 & .3950 & .4120 (32) & .4490 (8) & 1.000 (97185) \\
    Agnews       & .2500 & .4950 & .6410 (32) & .6900 (2) & .8310 (49369) \\
    Amazon       & .2000 & .3750 & .2750 (256) & .4680 (128) & 1.000 (85696) \\
    IMDB         & .5000 & .9660 & .9630 (8192) & .9650 (1024) & .9710 (15892) \\
    MyriadLAMA   & .2500 & .5080 & .5200 (4) & .5760 (4) & 1.000 (80167) \\
    FEVER        & .5000 & .6530 & .7830 (32) & .7610 (32) & .7920 (45564) \\
    CSQA         & .2000 & .5170 & .3490 (1) & .5380 (16) & .5730 (96696) \\
    TempLAMA     & .2500 & .2430 & .3560 (4096) & .3640 (16) & 1.000 (113786) \\
    PAWS         & .5000 & .5000 & .7640 (128) & .7920 (128) & 1.000 (58200) \\
    MNLI         & .3333 & .3560 & .4980 (4) & .5590 (128) & .6740 (79711) \\
    SWAG         & .2500 & .4610 & .3360 (512) & .5310 (2) & .5160 (96955) \\
    HaluEval     & .5000 & .4990 & .7540 (1024) & .7510 (32) & 1.000 (58987) \\
    Toxic        & .5000 & .7230 & .8250 (1024) & .8210 (16) & .8390 (32263) \\
    Stereoset    & .3333 & .1096 & .8299 (16) & .7335 (16) & .8847 (29242) \\
    M-Amazon     & .2000 & .2990 & .2350 (4096) & .3740 (2) & .6260 (97623) \\
    LTI          & .2000 & .3670 & .4300 (4) & .5830 (8) & .9970 (12068) \\
    mLAMA        & .2500 & .4020 & .3880 (128) & .4470 (4) & .4640 (79839) \\
    XNLI         & .3333 & .3270 & .3500 (256) & .3620 (16) & .4510 (79212) \\
    M-POS        & .2500 & .3890 & .2610 (1024) & .3930 (4) & .7740 (90001) \\
    \bottomrule
    \end{tabular}
    }
    \caption{Optimal accuracies across all MCEval8K tasks in the zero-shot setting on Llama2-7B, with the neuron sizes achieving these accuracies.}
    \label{tab:per-task-acc-zeroshot}
\end{table*}

\begin{table*}[t]
    \centering
    \normalsize
    \setlength{\tabcolsep}{5.5pt}
    \resizebox{0.8\textwidth}{!}{
    \begin{tabular}{llllll}
    \toprule
    Tasks & \textbf{\makecell[c]{Rand}} & \textbf{\makecell[c]{LM-Prob}} & \textbf{
\makecell[c]{Polar-Probe\\(\#n\_neurons)}} & \textbf{\makecell[c]{Magn-Probe\\(\#n\_neurons)}} & \textbf{\makecell[c]{Tree-Probe\\(\#n\_neurons)}} \\ \midrule
    GED          & .5000 & .5060 & .8330 (16) & .8330 (64) & 1.000 (43465) \\
    POS          & .2500 & .5730 & .5870 (4) & .6210 (16) & .6550 (80695) \\
    CHUNK        & .2500 & .2710 & .2820 (8192) & .3910 (64) & 1.000 (101539) \\
    NER          & .2500 & .3610 & .4300 (4) & .4970 (64) & 1.000 (93577) \\
    Agnews       & .2500 & .5880 & .7060 (64) & .6890 (512) & .8120 (42846) \\
    Amazon       & .2000 & .4840 & .5310 (1) & .5680 (128) & 1.000 (84055) \\
    IMDB         & .5000 & .9700 & .9700 (64) & .9690 (64) & .9660 (13823) \\
    MyriadLAMA   & .2500 & .7380 & .7450 (256) & .7530 (4096) & .7460 (70446) \\
    FEVER        & .5000 & .6780 & .8000 (1) & .8030 (4) & .8210 (38943) \\
    CSQA         & .2000 & .6100 & .6180 (32) & .6340 (8192) & .6180 (94246) \\
    TempLAMA     & .2500 & .2600 & .2500 (1) & .4110 (4) & 1.000 (106140) \\
    PAWS         & .5000 & .5240 & .8180 (16) & .8210 (32) & 1.000 (44060) \\
    MNLI         & .3333 & .5100 & .5780 (32) & .5860 (64) & .6830 (67771) \\
    SWAG         & .2500 & .4100 & .4430 (256) & .4710 (64) & .4160 (95311) \\
    HaluEval     & .5000 & .5200 & .7750 (2048) & .7770 (256) & 1.000 (51411) \\
    Toxic        & .5000 & .7800 & .8250 (8) & .8260 (4) & .8430 (29766) \\
    Stereoset    & .3333 & .1040 & .7297 (128) & .5180 (16) & .8204 (29774) \\
    M-Amazon     & .2000 & .5250 & .5470 (1024) & .5880 (128) & .6820 (87424) \\
    LTI          & .2000 & .3680 & .5480 (64) & .6950 (8) & .9910 (28362) \\
    mLAMA        & .2500 & .6080 & .6230 (8192) & .6360 (512) & .6450 (75439) \\
    XNLI         & .3333 & .3970 & .4860 (32) & .4980 (32) & .5990 (80886) \\
    M-POS        & .2500 & .4440 & .4830 (4) & .5130 (8) & 1.000 (95537) \\
    \bottomrule
    \end{tabular}
    }
    \caption{Optimal accuracies across all MCEval8K tasks in the few-shot setting on Llama2-7B, with the neuron sizes achieving these accuracies. The number of demonstrations is set to the same number of options for each task.}
    \label{tab:per-task-acc-fewshot}
\end{table*}

\subsection{Random Forest-based Probe}
\label{sec:appendix-random-forest-probe}

\smallskip\noindent\textbf{What is the random forest algorithm?} 
The random forest is an ensemble learning algorithm~\citep{randomforest} that works by creating a multitude of decision trees during training.
For our multi-choice classification tasks in MCEval8K, the random forest's output is the option selected by most trees.
A decision tree is a supervised learning model that makes predictions by recursively splitting data based on feature values. 
During training, the tree builds nodes by selecting features that best separate the data according to a chosen metric, such as Gini impurity. 
Splitting continues until the data in each leaf node is sufficiently pure or a maximum depth is reached. 
During inference, a new input is passed through the tree by following the feature-based decisions from the root to a leaf, where the final prediction is determined by the majority label or average value of samples in that leaf.

\smallskip\noindent\textbf{Feature design:} 
Our study's objective is to explore the effectiveness of using NEG as features for knowledge representation and conduct further analysis. 
Therefore, the inputs for training and inference in the random forest model are constructed solely based on gradients estimated by NeurGrad. 
Specifically, each neuron is assigned an integer value for a given prompt. 
In our classification tasks, a neuron’s feature is set to $i$ if the gradient associated with the $i$-th token for that neuron is the largest among the gradients computed for all other candidate tokens (options).
We ignore information on the smallest gradients to reduce the size of feature spaces.

\smallskip\noindent\textbf{Implementation details:}
For the implementation, we directly use \textit{RandomForestClassifier} in scikit-learn~\cite{scikit-learn} for training and inference.
We use the default parameters of \href{https://scikit-learn.org/1.5/modules/generated/sklearn.ensemble.RandomForestClassifier.html#randomforestclassifier}{RandomForestClassifier} besides the number of trees (\#n\_trees) and layers (\#n\_layers) used in each tree.
The number of trees refers to the number of decision trees used to ensemble the random forest. 
The number of layers refers to the layer depth for each tree. 
Noted that RandomForestClassifier constructs binary trees; thus, the number of features used in each tree is equal to or less than $2^{\#\text{n\_layers}} - 1$.

\begin{figure*}[t]
    \centering
    \includegraphics[width=\linewidth]{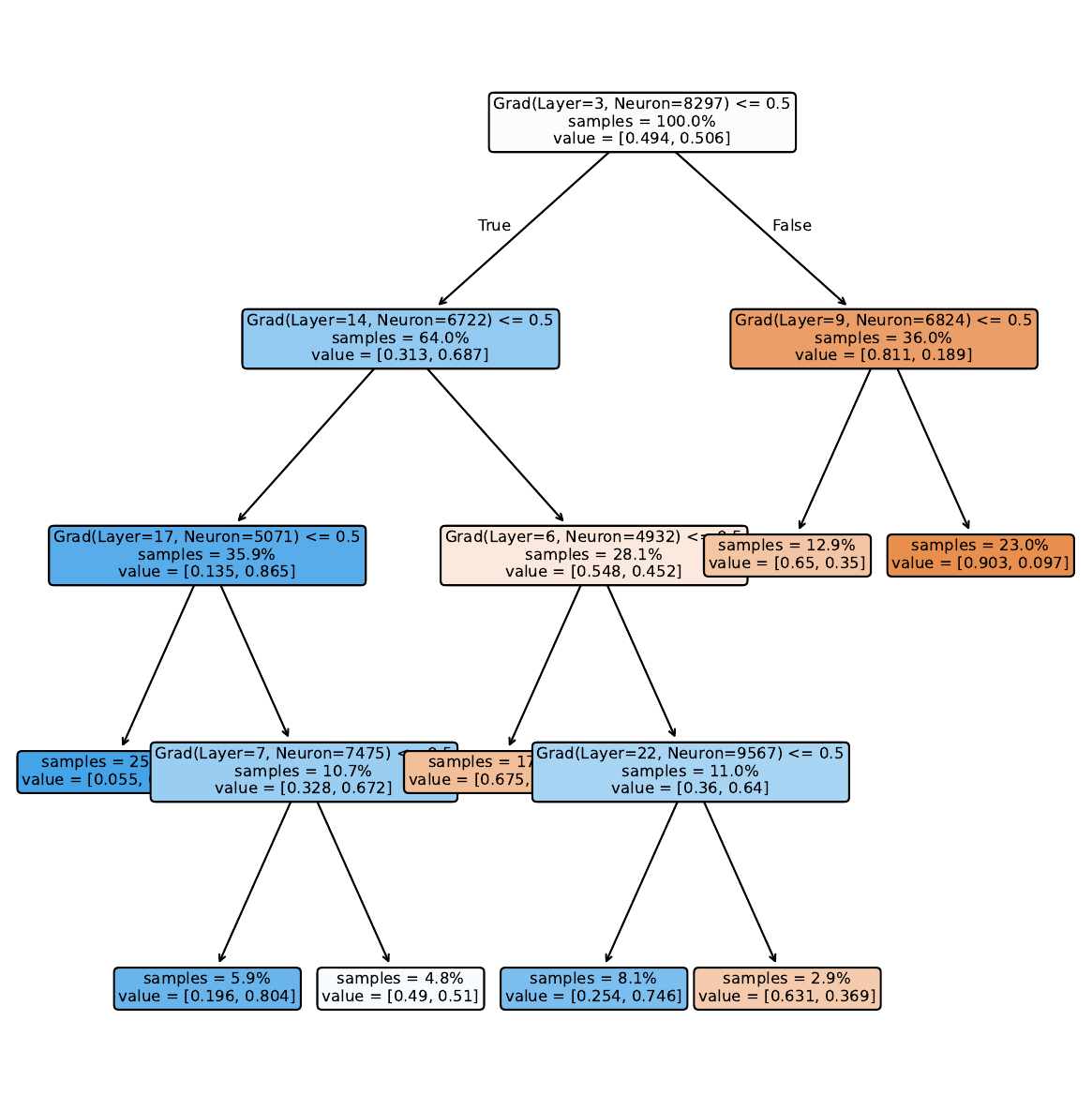}
    \caption{Visualization of a decision tree learned for the PAWS dataset with the few-shot setting on Llama2-7B model. The ``samples'' in each node refers to the percentage of samples reaching this node. The ``value'' shows the class distribution of samples in the node.}
    \label{fig:example-tree}
\end{figure*}

\smallskip\noindent\textbf{Visualization:}
We present an example of a single-tree random forest model learned from the PAWS dataset in the few-shot setting, illustrated in Figure~\ref{fig:example-tree}. 
For learning this decision tree, the number of trees and layers is set to 1 and 8. 
The PAWS dataset is a binary classification task with candidate tokens "yes" and "no." 
To construct features for each neuron, we compare the NEGs computed by NeurGrad for the "yes" and "no" prompt pairs. 
If the gradient estimated for the prompt-yes pair exceeds that of the prompt-no pair, we assign a feature value of 1; otherwise, we assign 0.

\section{Additional Analysis on Probing Results}
\label{sec:appendix-further-analysis}

% \subsection{Interpreting in-context learning with empirical gradients.} 
% \label{sec:appendix-interpreting-icl}
% To understand why simple majority-vote classifiers achieve high accuracy, we analyze the gradients associated with each answer choice. 
% Using PAWS (binary classification) as an example, we inspect the gradient pairs for target tokens (yes/no) across all training prompts. 
% We find that 97.21\% of neurons display opposite signs for yes/no tokens. 
% Moreover, the Corr between yes/no gradients is -0.9996. This pronounced inverse Corr suggests that empirical gradients are sharply polarized, making it easier for a majority-vote approach to distinguish between the target tokens.
% Furthermore, we examine how zero-shot and few-shot prompting differ from the perspective of empirical gradients. 
% Our analysis reveals that the total gradient magnitudes in few-shot scenarios over 22 tasks are 5.36 times greater than in zero-shot. 
% This indicates that demonstrations in context can effectively activate skill neurons, leading to better task understanding. 

\subsection{More Data about Efficiency}
\label{sec:appendix-training-efficiency}

\begin{figure*}[t]
    \centering
    \includegraphics[width=\linewidth]{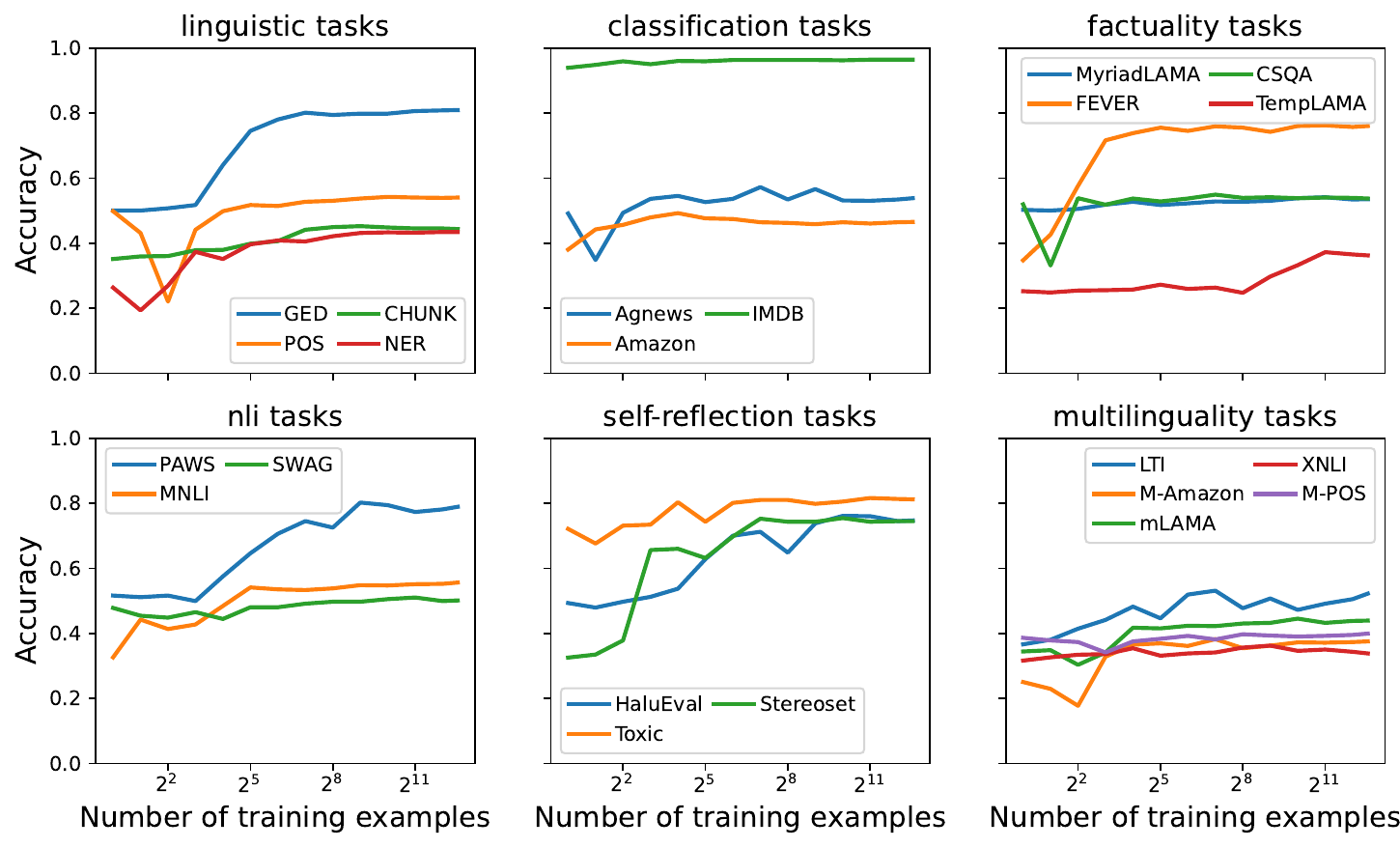}
    \caption{Per-task accuracies with the varying number of training examples on Llama2-7B, zero-shot prompt setting.}
    \label{fig:trainsize2accuracy-pertasks-zero}
\end{figure*}

\begin{figure*}[t]
    \centering
    \includegraphics[width=\linewidth]{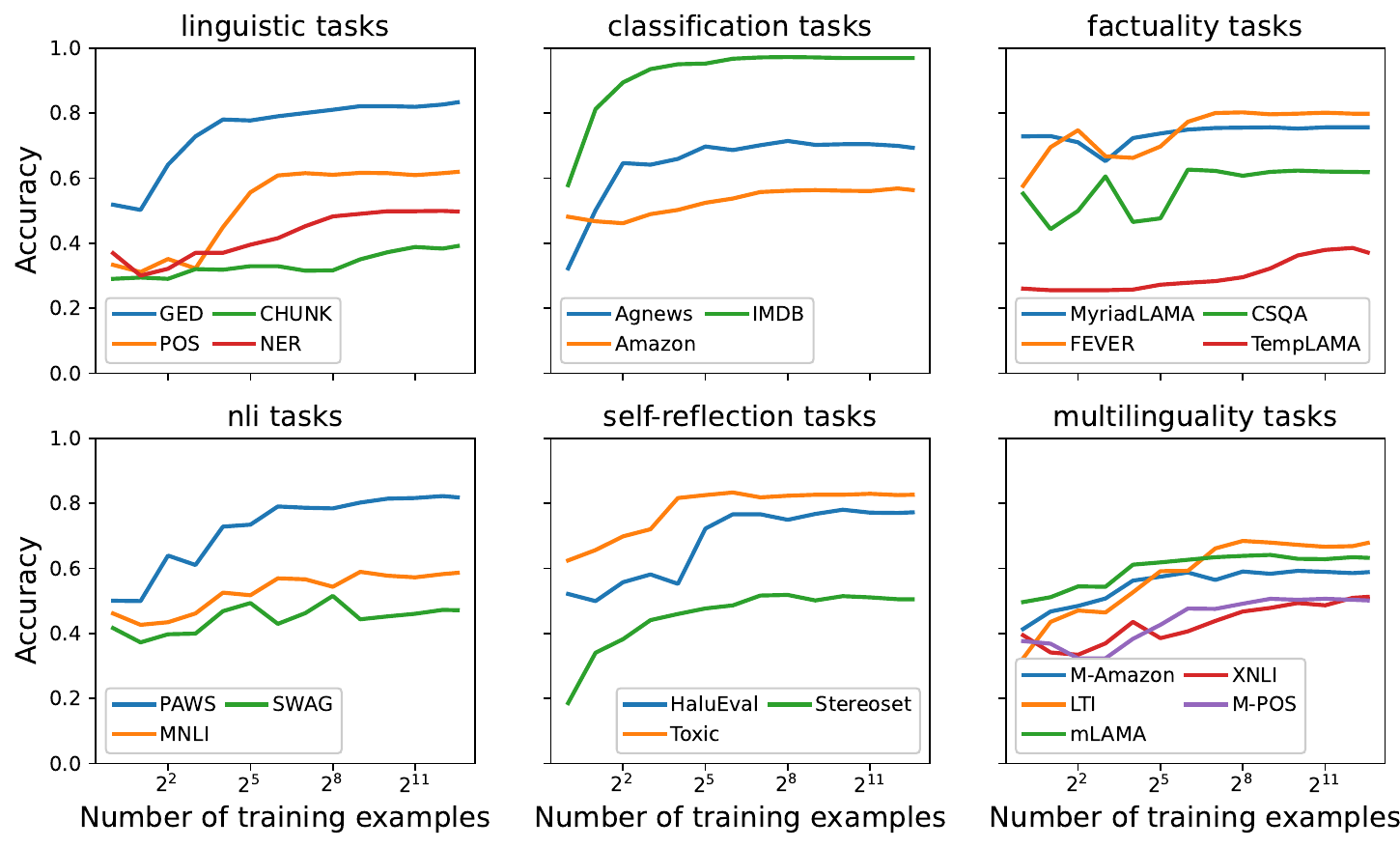}
    \caption{Per-task accuracies with the varying number of training examples on Llama2-7B, few-shot prompt setting.}
    \label{fig:trainsize2accuracy-pertasks}
\end{figure*}

We report the accuracies of majority-vote probes with different neuron sizes for all tasks to provide additional evidence for the discussion about the representation and acquisition efficiency of skill neurons in \S~\ref{sec:neuron-efficiency}.
The results are demonstrated in Figure~\ref{fig:trainsize2accuracy-pertasks-zero} and Figure~\ref{fig:trainsize2accuracy-pertasks} for zero-shot and few-shot prompting settings. 

\begin{figure*}[t]
    \centering
    \includegraphics[width=\linewidth]{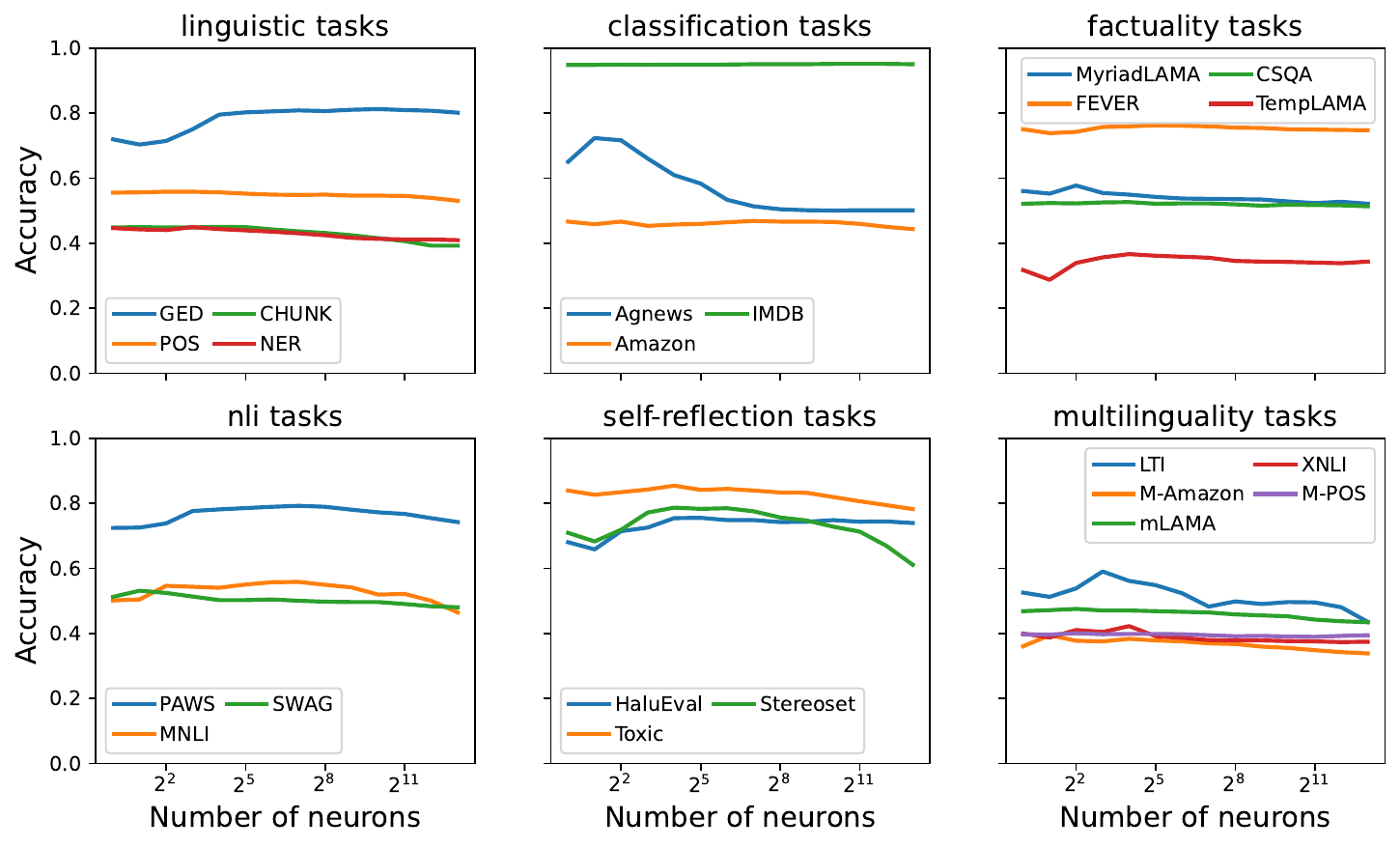}
    \caption{Per-task accuracies with varying neuron sizes on Llama2-7B, zero-shot prompt setting.}
    \label{fig:neurinsize2accuracy-pertasks-zeroshot}
\end{figure*}

\begin{figure*}[t]
    \centering
    \includegraphics[width=\linewidth]{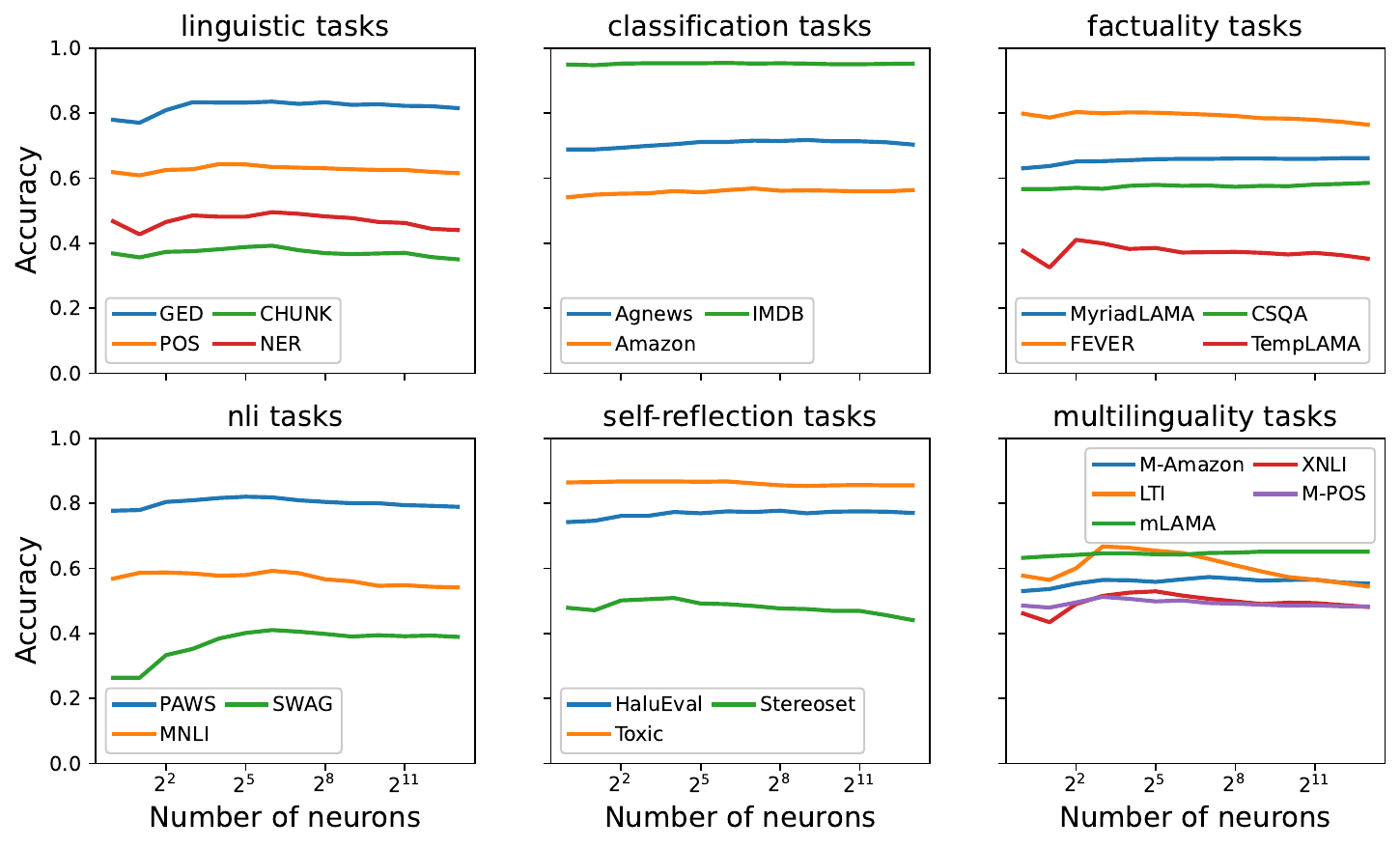}
    \caption{Per-task accuracies with varying neuron sizes on Llama2-7B, few-shot prompt setting.}
    \label{fig:neurinsize2accuracy-pertasks}
\end{figure*}

\begin{figure*}[t]
    \centering
    \includegraphics[width=\linewidth]{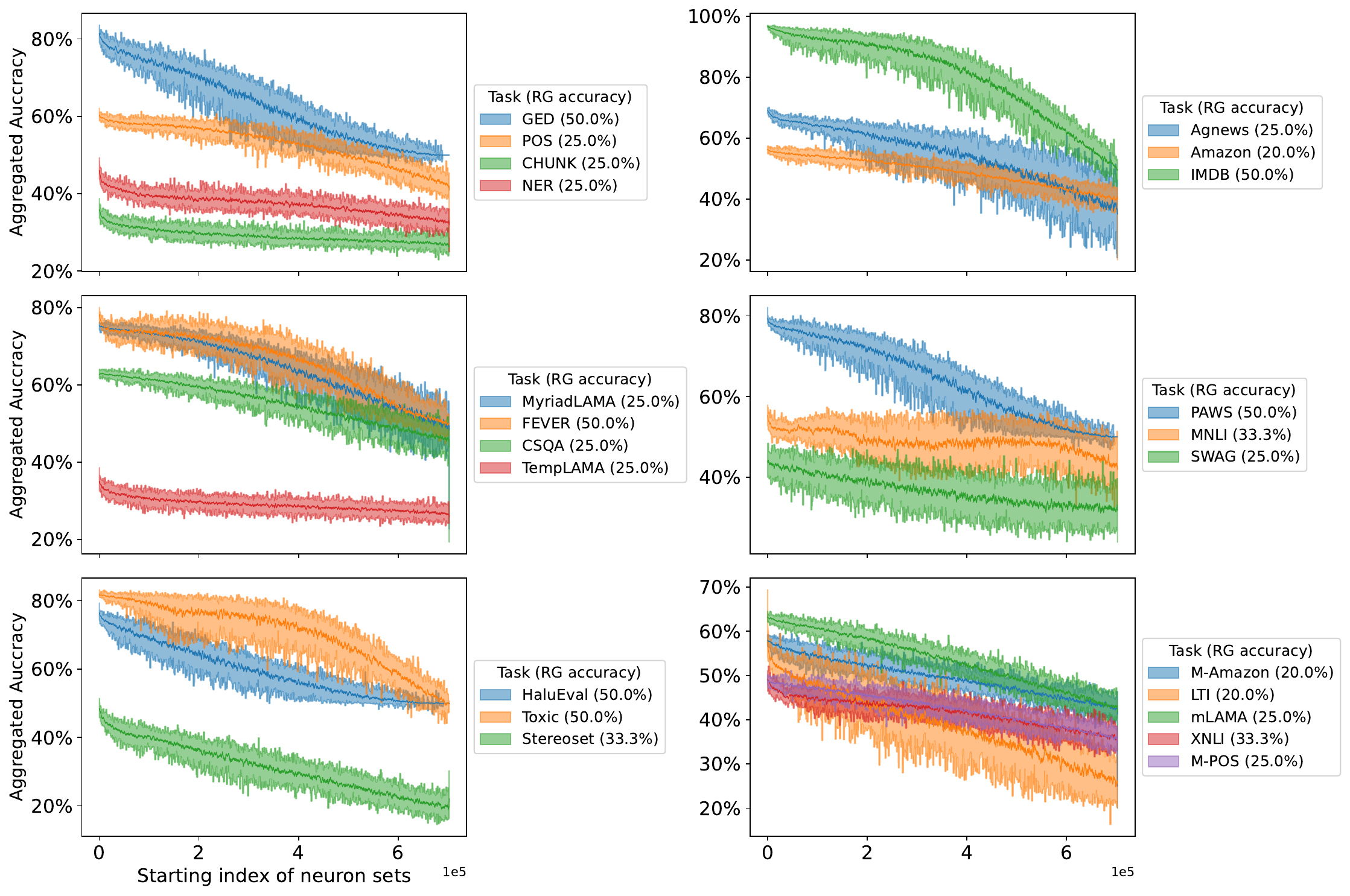}
    \caption{Per-task accuracies with varying neuron sets per with 64 neurons. We report the aggregated accuracies with a window size of 64 for better visualization, plotting the mean accuracy within each window, along with the corresponding accuracy ranges (minimum to maximum) as the envelope.}
    \label{fig:full-inclusivity}
\end{figure*}

\begin{figure*}[t]
    \centering
    \includegraphics[width=0.7\linewidth]{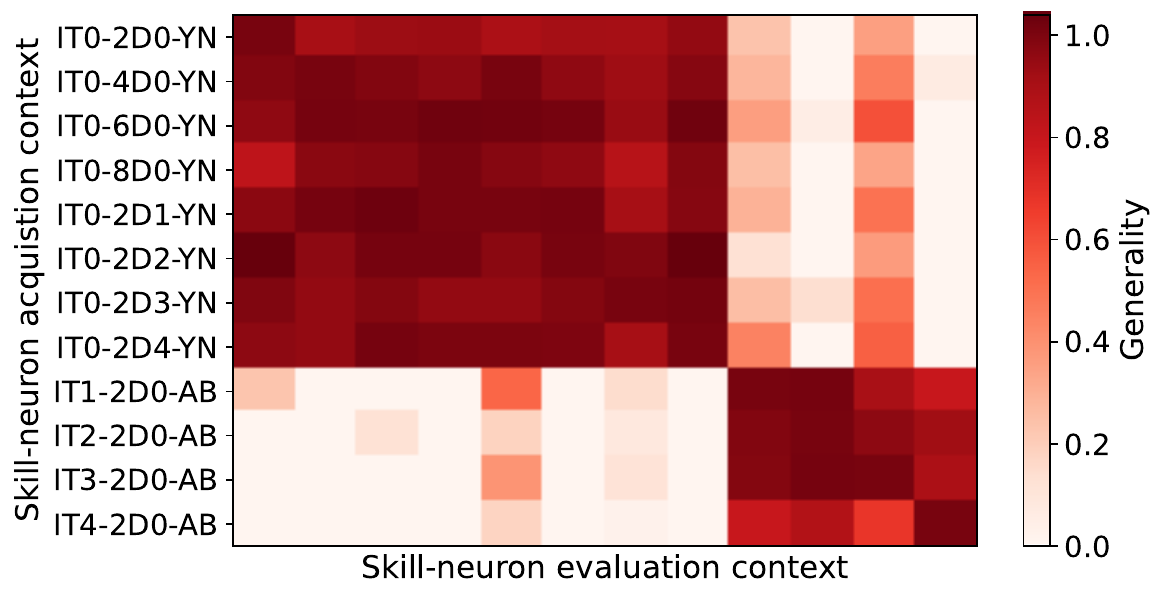}
    \caption{Generality of skill neurons across different contexts. \textbf{X-axis}: the context used to acquire skill neurons. \textbf{Y-axis}: evaluation context. The contexts on the x-axis are in the same order as on the y-axis.
    The context using the i-th instruction, k-th set of j-shot demonstrations, and yes/no answers is denoted as IT(i)-(j)D(k)-YN\@. ``AB'' refers to the a/b style options.
    }
    \label{fig:generality}
\end{figure*}

\subsection{Probing With Varying Neuron Sets}
\label{sec:appendix-neuron-set}

We report the aggregated accuracies across all 22 tasks in MCEval8K in Figure~\ref{fig:full-inclusivity} to provide additional evidence for discussion in \S~\ref{sec:neuron-inclusivity}. 
It demonstrates that many neurons can construct the classifiers in solving the language tasks, showing their ability to represent language skills and knowledge.

\subsection{Tree-Probe: Flatteness vs. Hierarchy}
\label{sec:appendix-neuron-hierarchy}
% Finally, we ask: do skill neurons depend on each other? 
% The major-vote probe assumes independence between neurons, while the Tree-Probe models their interdependencies and builds hierarchical classifiers. 
% Its significant advantage over the major-vote probe in Table~\ref{tab:average-acc} suggests that language skills are hierarchically represented. 
% However, it remains unclear how much hierarchy is needed.

To investigate the balance between hierarchy and independence of skill neurons, we train Tree-Probe with fixed neuron features ($2^{10}$) but with different depths and trees.
For each task, we train 10 Tree-Probes, varying the number of trees (\#n\_tree $\in (2^{0} \sim 2^{9})$) and the tree depth (\#n\_layer $\in (2^{10} \sim 2^{1})$), which fewer trees with deeper layers indicate a more hierarchical structure. 
All tasks show a camel curve given stronger hierarchies.
We report the optimal \#n\_layer for different tasks as follows: CSQA(4), MNLI(16), SWAG(16), Stereoset(16), Agnews(32), MyriadLAMA(32), mLAMA(32), XNLI(32), POS(64), FEVER(64), Toxic(64), LTI(64), GED(128), IMDB(128), M-Amazon(128), CHUNK(256), NER(256), Amazon(256), PAWS(256), HaluEval(256), M-POS(256), TempLAMA(1024). 
This demonstrates that Different language skills require different hierarchy levels. For instance, factual tasks benefit from flatter structures, while linguistic tasks prefer deeper hierarchies. 

\begin{figure*}[t]
    \centering
    \includegraphics[width=0.97\linewidth]{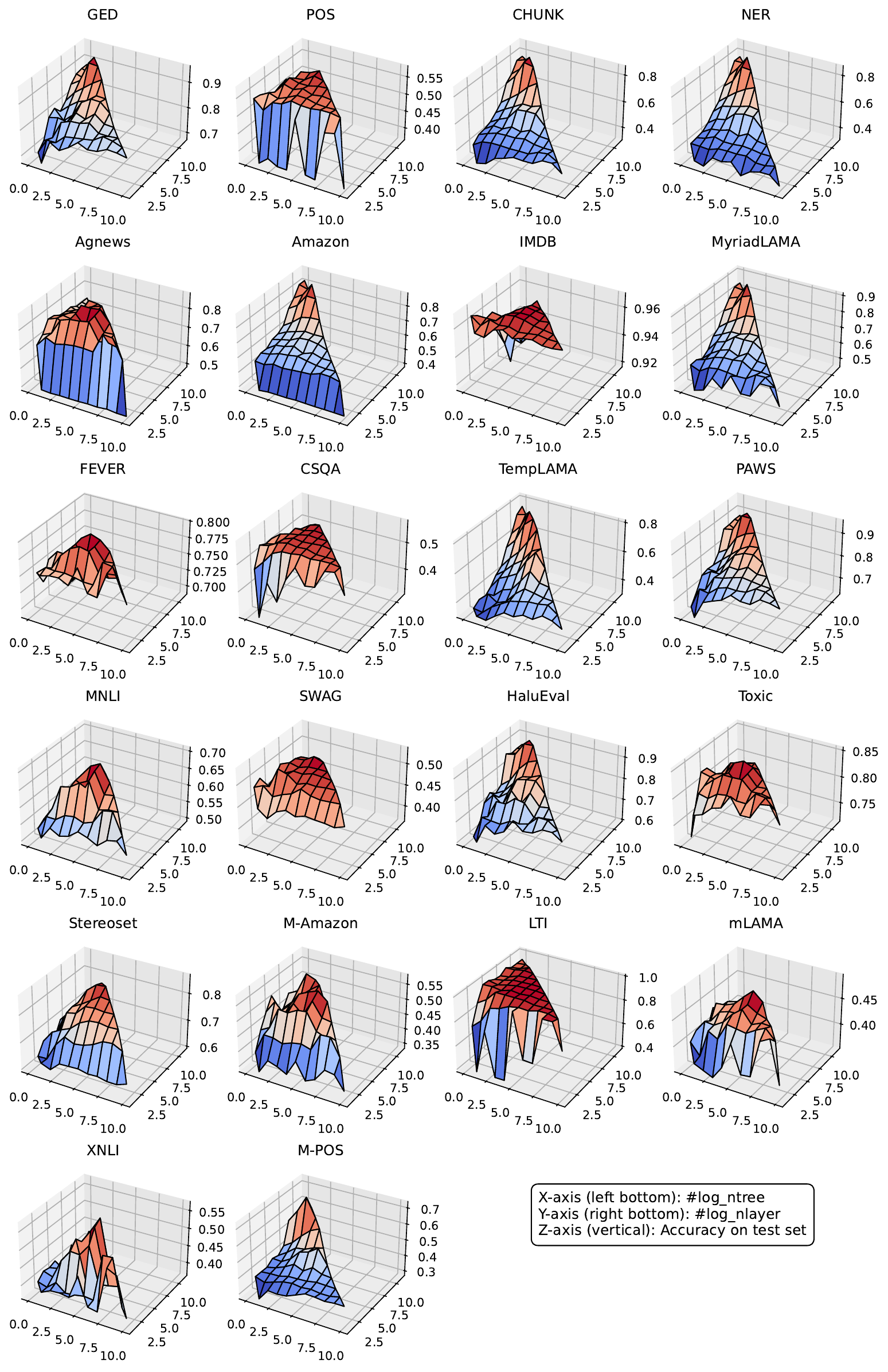}
    \caption{Accuracies of trained random forest models with the zero-shot setting on Llama2-7B. }
    \label{fig:tree-full-accs-zero-shot}
\end{figure*}

\begin{figure*}[t]
    \centering
    \includegraphics[width=0.97\linewidth]{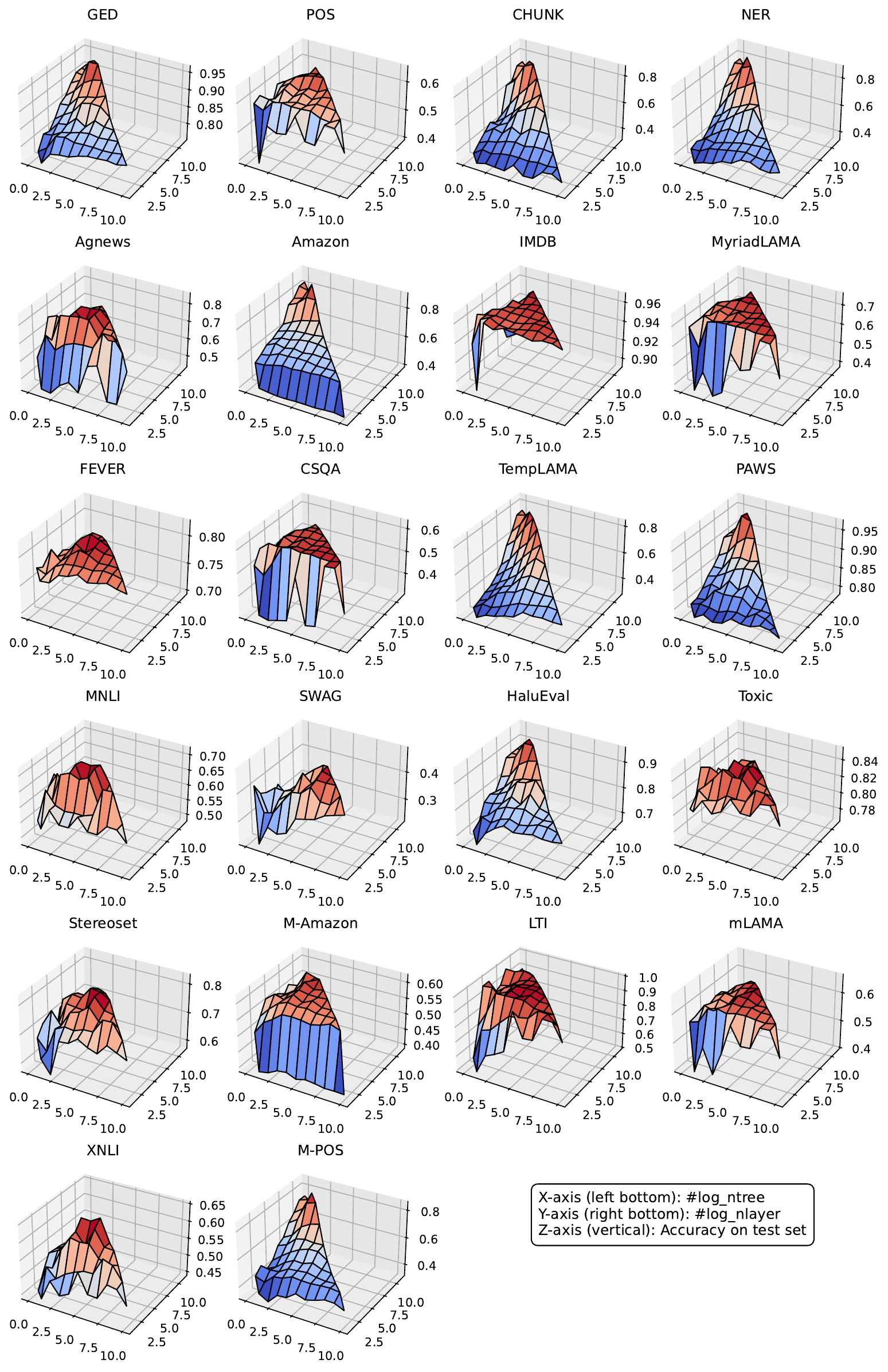}
    \caption{Accuracies of trained random forest models with the few-shot setting on Llama2-7B. }
    \label{fig:tree-full-accs-few-shot}
\end{figure*}

For all tasks in MCEval8K, we plot the accuracies of trained models with varying hyperparameters, including the number of trees and layers per tree.
The number of trees is set to $2^N$, where $N$ ranges from 0 to 10, and the number of layers is set to $2^M$, where $M$ ranges from 1 to 11.
Training is conducted only for configurations where $N + M < 12$. 
The results are visualized as 3D surfaces, where the x-axis represents the logarithm of the number of trees (\#log\_ntree), the y-axis shows the logarithm of the number of layers (\#log\_nlayer), and the z-axis indicates the accuracy evaluated on the test set.
We display the results for all tasks under the zero-shot setting in Figure~\ref{fig:tree-full-accs-zero-shot} and those under the few-shot setting in Figure~\ref{fig:tree-full-accs-few-shot}. 

% \section{Activation-based vs. Gradient-based Skill Neurons}
% \label{sec:appendix-activation-skill-neurons}

\section{Prompting Setups}
\label{sec:appendix-prompt-instructions}

In this subsection, we list all the instructions we use for each task in MCEval8K. 
It includes design instructions, options, and a selection of few-shot examples. 
As mentioned in \S~\ref{subsec:probing_experiment_setp}, we adopt two instruction settings, zero-shot and few-shot. 
For few-shot prompting, we set the number of examples to the same number as the number of options and ensure each option only appears once to prevent majority label bias~\cite{zhao2021calibrateuseimprovingfewshot}.
All the few-shot examples are sampled from the training set. 
Finally, we list all the instructions and options we used for skill neuron probing examples by showing one zero-shot prompt.

\paragraph{GED}

\texttt{
\newline
\noindent\#\#\# Instruction: Which of the sentence below is linguistically acceptable?\newline
\noindent\#\#\# Sentences:\newline
\noindent a.I set the alarm for 10:00 PM but I could n't wake up then .\newline
\noindent b.I set the alarm for 10:00PM but I could n't wake up then .\newline
\noindent \#\#\# Answer:}

\paragraph{POS}
\texttt{
\newline
\noindent\#\#\# Instruction: Determine the part-of-speech (POS) tag for the highlighted target word in the given text. Choose the correct tag from the provided options.\newline
\noindent\#\#\# Input text:One of the largest population centers in pre-Columbian America and home to more than 100,000 people at its height in about 500 CE, Teotihuacan was located about thirty miles northeast of modern Mexico City.\newline
\#\#\# Target word:'pre-Columbian'\newline
\#\#\# Options:\newline
a.DET\newline
b.ADJ\newline
c.PRON\newline
d.PUNCT\newline
\#\#\# Answer:
}

\paragraph{CHUNK}
\texttt{
\newline
\noindent\#\#\# Instruction: Identify the chunk type for the specified target phrase in the sentence and select the correct label from the provided options.\newline
\noindent\#\#\# Input text:B.A.T said it purchased 2.5 million shares at 785 .\newline
\noindent\#\#\# Target phrase:'said'\newline
\noindent\#\#\# Options:\newline
a.PP\newline
b.VP\newline
c.NP\newline
d.ADVP\newline
\noindent\#\#\# Answer:
}

\paragraph{NER}
\texttt{
\newline
\noindent\#\#\# Identify the named entity type for the specified target phrase in the given text. Choose the correct type from the provided options\newline
\noindent\#\#\# Input text:With one out in the fifth Ken Griffey Jr and Edgar Martinez stroked back-to-back singles off Orioles starter Rocky Coppinger ( 7-5 ) and Jay Buhner walked .\newline
\noindent\#\#\# Target phrase:'Orioles'\newline
\noindent\#\#\# Options:\newline
a.LOC\newline
b.ORG\newline
c.MISC\newline
d.PER\newline
\noindent\#\#\# Answer:}

\paragraph{Agnews}
\texttt{
\newline
\noindent\#\#\# Instruction: Determine the genre of the news article. Please choose from the following options: a.World b.Sports c.Business d.science. Select the letter corresponding to the most appropriate genre.\newline
\noindent\#\#\# Text:Context Specific Mirroring \\"Now, its not that I dont want to have this content here. Far from it. Ill\\always post everything to somewhere on this site. I just want to treat each\\individual posting as a single entity and place it in as fertile a set of beds\\as possible. I want context specific mirroring. I want to be able to \\newlinechoose\\multiple endpoints for a post, and publish to all of them with a single button\\click."\\\newline
\noindent\#\#\# Genres:\newline
a.World\newline
b.Sports\newline
c.Business\newline
d.Science\newline
\noindent\#\#\# Answer:
}

\paragraph{Amazon}
\texttt{
\newline
\noindent\#\#\# Instruction: Analyze the sentiment of the given Amazon review and assign a score from 1 (very negative) to 5 (very positive) based on the review. Output only the score.\newline
\noindent\#\#\# Input Review:I never write reviews, but this one really works, doesn't float up, is clean and fun. Kids can finally take a bath!\newline
\noindent\#\#\# Output Score:
}

\paragraph{IMDB}
\texttt{
\newline
\noindent\#\#\# Instruction: Based the review, is the movie good or bad?\newline
\noindent\#\#\# Review:Stewart is a Wyoming cattleman who dreams to make enough money to buy a small ranch in Utah ranch <...abbreviation...>. In spontaneous manner, Stewart is lost between the ostentatious saloon owner and the wife-candidate...\newline
\noindent\#\#\# Answer:
}

\paragraph{MyriadLAMA}
\texttt{
\newline
\noindent\#\#\# Instruction: Predict the [MASK] in the sentence from the options. Do not provide any additional information or explanation.\newline
\noindent\#\#\# Question:What is the native language of Bernard Tapie? [MASK].\newline
\noindent\#\#\# Options:\newline
a.Dutch\newline
b.Telugu\newline
c.Russian\newline
d.French\newline
\noindent\#\#\# Answer:
}

\paragraph{CSQA}
\texttt{
\newline
\noindent\#\#\# Instruction: Please select the most accurate and relevant answer based on the context.\newline
\noindent\#\#\# Context: What does a lead for a journalist lead to?\newline
\noindent\#\#\# Options:\newline
a.very heavy\newline
b.lead pencil\newline
c.store\newline
d.card game\newline
e.news article\newline
\noindent\#\#\# Answer:
}

\paragraph{TempLAMA}
\texttt{
\newline
\noindent\#\#\# Instruction: Select the correct year from the provided options that match the temporal fact in the sentence. Output the index of the correct year.\newline
\noindent\#\#\# Question:Pete Hoekstra holds the position of United States representative.\newline
\noindent\#\#\# Options:\newline
a.2013\newline
b.2014\newline
c.2018\newline
d.2011\newline
\noindent\#\#\# Answer:
}

\paragraph{PAWS}
\texttt{
\newline
\noindent\#\#\# Instruction: Is the second sentence a paraphrase of the first? Answer exactly 'yes' or 'no'.\newline
\noindent\#\#\# Sentence 1: It is directed by Kamala Lopez and produced by Cameron Crain , Richard Shelgren and Kamala Lopez .\newline
\noindent\#\#\# Sentence 2: It was produced by Cameron Crain , Richard Shelgren and Kamala Lopez and directed by Kamala Lopez .\newline
\noindent\#\#\# Answer:
}

\paragraph{MNLI}
\texttt{
\newline
\noindent\#\#\# Instruction: Given a premise and a hypothesis, determine the relationship.\newline
\noindent\#\#\# Premise: easily yeah yeah and then if you want popcorn and stuff it's just i mean uh it's incredible\newline
\noindent\#\#\# Hypothesis: It's anti-incredible, very ordinary and unimpressive.\newline
\noindent\#\#\# Question: What is the relationship between the two sentences?.\newline
\noindent\#\#\# Options:\newline
a.Entailment\newline
b.Neutral\newline
c.Contradiction\newline
\noindent\#\#\# Answer:
}

\paragraph{SWAG}
\texttt{
\newline
\noindent\#\#\# Instruction: Given the context, select the most likely completion from the following choices. Please exactly answer the label.\newline
\noindent\#\#\# Context: He looks back at her kindly and watches them go. In someone's dark bedroom, someone\newline
\noindent\#\#\# Options:\newline
a.paces with the bandage, his back to someone.\newline
b.spies a framed photo of a burmese soldier on a black horse.\newline
c.blinks covers the apartment's couch.\newline
d.lays her sleeping niece down gently onto the bed.\newline
\noindent\#\#\# Answer:
}

\paragraph{HaluEval}
\texttt{
\newline
\noindent\#\#\# Instruction: Given the knowledge context, dialogue histroy and response, determine if any hallucination is present. Provide a response of either 'yes' or 'no' only.\newline
\noindent\#\#\# Context:Kim Edwards wrote The Memory Keeper's Daughter\newline
\noindent\#\#\# Dialogue history:[Human]: Could you recommend something by Kim Edwards? [Assistant]: Absolutely, The Memory Keeper's Daughter is good. Do you like Fiction? [Human]: Yes, I do. I really love Sci Fi. [Assistant]: OK. Some Sci Fi choices are Blade Runner, The Matrix and Alien. Do you like any of these? [Human]: I wasn't fond of Alien or Blade Runner but I loved The Matrix. \newline
\noindent\#\#\# Response:Actually, Kim Edwards is a director and her best work so far is The Memory Keeper's Daughter. For Sci-Fi, I suggest you watch the 1989 classic Field of Dreams. Would you like to give it a try?\newline
\noindent\#\#\# Hallucination (yes or no):
}

\paragraph{Toxic}
\texttt{
\newline
\noindent\#\#\# Instruction: Determine if the provided text contains toxic content.Provide a response of either 'yes' or 'no' only.\newline
\noindent\#\#\# Text:Dear SchfiftyThree\newline
I would just like to say that i hate you and the rest of the wikipedia editors for deleting my creations. fuck you.\newline
\noindent\#\#\# Toxicity (yes or no):
}

\paragraph{Stereoset}
\texttt{
\newline
\noindent\#\#\# Instruction: Given the context, identify and select the sentence that does not convey the stereotype related to the context.\newline
\noindent\#\#\# Context:A performing artist is often up late at night.\newline
\noindent\#\#\# Options:\newline
a.Many artists are very likeable people.\newline
b.Having children is a momentous occasion.\newline
c.Any who performs is a diva.\newline
\noindent\#\#\# Answer:
}

\paragraph{M-Amazon}
\texttt{
\newline
\noindent\#\#\# Anweisung: Analysieren Sie die Stimmung der gegebenen Amazon-Bewertung und vergeben Sie eine Punktzahl von 1 (sehr negativ) bis.\newline
\noindent\#\#\# Eingabebewertung:Produkt ist gefährlich. Hat den Fi rausgehen. Man hat das gefühl es fällt auseinander. Billigste Qualität!! Man fühlt sich einfach betrogen!!!\newline
\noindent\#\#\# Ausgabewertung:
}

\paragraph{LTI}
\texttt{
\newline
\noindent\#\#\# Instruction: Identify the language of the given sentence.\newline
\noindent\#\#\# Text:S'en retournait, et assis sur son chariot, lisait le prophète Ésaïe.\newline
\noindent\#\#\# Options:\newline
a.English\newline
b.French\newline
a.German\newline
a.Chinese\newline
a.Spanish\newline
\noindent\#\#\# Answer:
}

\paragraph{mLAMA}
\texttt{
\newline
\noindent\#\#\# Instrucción: Prediga el [MASK] en la oración a partir de las opciones. No proporcione información ni explicaciones adicionales.\newline
\noindent\#\#\# Respuesta:La capital de Irán es [MASK].\newline
\noindent\#\#\# Opciones:\newline
a.Indianápolis\newline
b.Génova\newline
c.Teherán\newline
d.París\newline
\noindent\#\#\# Pregunta:
}

\paragraph{XNLI}
\texttt{
\newline
\noindent\#\#\# Instruction: Étant donné une prémisse et une hypothèse, déterminez la relation.\newline
\noindent\#\#\# Prémisse: Ouais nous sommes à environ km au sud du lac Ontario en fait celui qui a construit la ville était un idiot à mon avis parce qu' ils l' ont construit ils l' ont construit assez loin de la ville qu' il ne pouvait pas être une ville portuaire\newline
\noindent\#\#\# Hypothèse: Nous sommes à 10 km au sud du lac Ontario en bas i-35 .\newline
\noindent\#\#\# Options:\newline
a.Implication\newline
b.Neutre\newline
c.Contradiction\newline
\noindent\#\#\# Réponse:
}

\paragraph{M-POS}
\texttt{
\newline
\begin{CJK}{UTF8}{gbsn}
\noindent\#\#\# 指令：确定给定文本中高亮目标词的词性。从提供的选项中选择正确的词性标签。\newline
\noindent\#\#\# 文本:但是，有一\begin{CJK}{UTF8}{bsmi}個\end{CJK}全面的人口\begin{CJK}{UTF8}{bsmi}統計數據\end{CJK}分析，\begin{CJK}{UTF8}{bsmi}對象\end{CJK}包括\begin{CJK}{UTF8}{bsmi}婦女\end{CJK}，特別是有\begin{CJK}{UTF8}{bsmi}養\end{CJK}育孩子的那些。\newline
\noindent\#\#\# 目标词:'一'\newline
\noindent\#\#\# 选项:\newline
a.NUM\newline
b.AUX\newline
c.ADJ\newline
d.VERB\newline
\noindent\#\#\# 问题:
\end{CJK}
}

\section{Diverse Contexts for Skill Neuron Generality Evaluation}
\label{sec:appendix-generality-instructions}

In this section, we report the instructions we used for experiments to measure the generality of skill neurons in \S~\ref{sec:neuron-generality}.
We report five types of instruction settings with 2-shot, $\text{IT0}, \text{IT1}, \text{IT2}, \text{IT3}, \text{IT4}$, where $\text{IT0}$ use yes/no as it candidate target tokens while others use a/b. 

We fix the number of skill neurons to 32 when training the skill-neuron-based probes. We use 32 as the optimal neuron size of PAWS with the few-shot setting is 32.
Finally, we report the pair-wise generality values among different prompting settings in Figure~\ref{fig:generality}. 

\paragraph{An example of $\text{IT0}$}

\texttt{
\newline
\noindent\#\#\# Instruction: Is the second sentence a paraphrase of the first? Answer exactly 'yes' or 'no'.\newline
\noindent\#\#\# Sentence 1: The canopy was destroyed in September 1938 by Hurricane New England in 1938 , and the station was damaged but repaired .\newline
\noindent\#\#\# Sentence 2: The canopy was destroyed in September 1938 by the New England Hurricane in 1938 , but the station was repaired .\newline
\noindent\#\#\# Answer:no\newline
\noindent\#\#\# Sentence 1: Pierre Bourdieu and Basil Bernstein explore , how the cultural capital of the legitimate classes has been viewed throughout history as the `` most dominant knowledge '' .\newline
\noindent\#\#\# Sentence 2: Pierre Bourdieu and Basil Bernstein explore how the cultural capital of the legitimate classes has been considered the `` dominant knowledge '' throughout history .\newline
\noindent\#\#\# Answer:yes\newline
\noindent\#\#\# Sentence 1: It is directed by Kamala Lopez and produced by Cameron Crain , Richard Shelgren and Kamala Lopez .\newline
\noindent\#\#\# Sentence 2: It was produced by Cameron Crain , Richard Shelgren and Kamala Lopez and directed by Kamala Lopez .\newline
\noindent\#\#\# Answer:
}

\paragraph{An example of $\text{IT1}$}

\texttt{
\newline
\noindent\#\#\# Instruction: Given two sentences, determine if they are paraphrases of each other.\newline
\noindent\#\#\# Sentence 1: The canopy was destroyed in September 1938 by Hurricane New England in 1938 , and the station was damaged but repaired .\newline
\noindent\#\#\# Sentence 2: The canopy was destroyed in September 1938 by the New England Hurricane in 1938 , but the station was repaired .\newline
\noindent\#\#\# Options:\newline
a.not paraphrase\newline
b.paraphrase\newline
\noindent\#\#\# Answer:a\newline
\noindent\#\#\# Sentence 1: Pierre Bourdieu and Basil Bernstein explore , how the cultural capital of the legitimate classes has been viewed throughout history as the `` most dominant knowledge '' .\newline
\noindent\#\#\# Sentence 2: Pierre Bourdieu and Basil Bernstein explore how the cultural capital of the legitimate classes has been considered the `` dominant knowledge '' throughout history .\newline
\noindent\#\#\# Options:\newline
a.not paraphrase\newline
b.paraphrase\newline
\noindent\#\#\# Answer:b\newline
\noindent\#\#\# Sentence 1: It is directed by Kamala Lopez and produced by Cameron Crain , Richard Shelgren and Kamala Lopez .\newline
\noindent\#\#\# Sentence 2: It was produced by Cameron Crain , Richard Shelgren and Kamala Lopez and directed by Kamala Lopez .\newline
\noindent\#\#\# Options:\newline
a.not paraphrase\newline
b.paraphrase\newline
\noindent\#\#\# Answer:
}

\paragraph{An example of $\text{IT2}$}

\texttt{
\newline
\noindent\#\#\# Instruction: Review the two given sentences and decide if they express the same idea in different words.\newline
\noindent\#\#\# Sentence 1: The canopy was destroyed in September 1938 by Hurricane New England in 1938 , and the station was damaged but repaired .\newline
\noindent\#\#\# Sentence 2: The canopy was destroyed in September 1938 by the New England Hurricane in 1938 , but the station was repaired .\newline
\noindent\#\#\# Options:\newline
a.non-equivalent\newline
b.equivalent\newline
\noindent\#\#\# Answer:a\newline
\noindent\#\#\# Sentence 1: Pierre Bourdieu and Basil Bernstein explore , how the cultural capital of the legitimate classes has been viewed throughout history as the `` most dominant knowledge '' .\newline
\noindent\#\#\# Sentence 2: Pierre Bourdieu and Basil Bernstein explore how the cultural capital of the legitimate classes has been considered the `` dominant knowledge '' throughout history .\newline
\noindent\#\#\# Options:\newline
a.non-equivalent\newline
b.equivalent\newline
\noindent\#\#\# Answer:b\newline
\noindent\#\#\# Sentence 1: It is directed by Kamala Lopez and produced by Cameron Crain , Richard Shelgren and Kamala Lopez .\newline
\noindent\#\#\# Sentence 2: It was produced by Cameron Crain , Richard Shelgren and Kamala Lopez and directed by Kamala Lopez .\newline
\noindent\#\#\# Options:\newline
a.non-equivalent\newline
b.equivalent\newline
\noindent\#\#\# Answer:\newline
}
\paragraph{An example of $\text{IT3}$}

\texttt{
\newline
\noindent\#\#\# Instruction: Examine the two sentences provided. Determine if the second sentence is a valid paraphrase of the first sentence.\newline
\noindent\#\#\# Sentence 1: The canopy was destroyed in September 1938 by Hurricane New England in 1938 , and the station was damaged but repaired .\newline
\noindent\#\#\# Sentence 2: The canopy was destroyed in September 1938 by the New England Hurricane in 1938 , but the station was repaired .\newline
\noindent\#\#\# Options:\newline
a.different\newline
b.similar\newline
\noindent\#\#\# Answer:a\newline
\noindent\#\#\# Sentence 1: Pierre Bourdieu and Basil Bernstein explore , how the cultural capital of the legitimate classes has been viewed throughout history as the `` most dominant knowledge '' .\newline
\noindent\#\#\# Sentence 2: Pierre Bourdieu and Basil Bernstein explore how the cultural capital of the legitimate classes has been considered the `` dominant knowledge '' throughout history .\newline
\noindent\#\#\# Options:\newline
a.different\newline
b.similar\newline
\noindent\#\#\# Answer:b\newline
\noindent\#\#\# Sentence 1: It is directed by Kamala Lopez and produced by Cameron Crain , Richard Shelgren and Kamala Lopez .\newline
\noindent\#\#\# Sentence 2: It was produced by Cameron Crain , Richard Shelgren and Kamala Lopez and directed by Kamala Lopez .\newline
\noindent\#\#\# Options:\newline
a.different\newline
b.similar\newline
\noindent\#\#\# Answer:\newline
}
\paragraph{An example of $\text{IT4}$}

\texttt{
\newline
\noindent\#\#\# Instruction: You are provided with two sentences. Identify whether they convey identical ideas or differ in meaning.\newline
\noindent\#\#\# Sentence 1: The canopy was destroyed in September 1938 by Hurricane New England in 1938 , and the station was damaged but repaired .\newline
\noindent\#\#\# Sentence 2: The canopy was destroyed in September 1938 by the New England Hurricane in 1938 , but the station was repaired .\newline
\noindent\#\#\# Options:\newline
a.The sentences convey different idea.\newline
b.The sentences convey the same ideas.\newline
\noindent\#\#\# Answer:a\newline
\noindent\#\#\# Sentence 1: Pierre Bourdieu and Basil Bernstein explore , how the cultural capital of the legitimate classes has been viewed throughout history as the `` most dominant knowledge '' .\newline
\noindent\#\#\# Sentence 2: Pierre Bourdieu and Basil Bernstein explore how the cultural capital of the legitimate classes has been considered the `` dominant knowledge '' throughout history .\newline
\noindent\#\#\# Options:\newline
a.The sentences convey different idea.\newline
b.The sentences convey the same ideas.\newline
\noindent\#\#\# Answer:b\newline
\noindent\#\#\# Sentence 1: It is directed by Kamala Lopez and produced by Cameron Crain , Richard Shelgren and Kamala Lopez .\newline
\noindent\#\#\# Sentence 2: It was produced by Cameron Crain , Richard Shelgren and Kamala Lopez and directed by Kamala Lopez .\newline
\noindent\#\#\# Options:\newline
a.The sentences convey different idea.\newline
b.The sentences convey the same ideas.\newline
\noindent\#\#\# Answer:\newline
}

\end{document}